\documentclass[journal,onecolumn]{IEEEtran}

\usepackage{lineno,hyperref}
\modulolinenumbers[5]
\usepackage{cite}
\usepackage{amsmath,amssymb,amsfonts}
\usepackage{graphicx}
\usepackage{textcomp}
\usepackage{multirow}
\usepackage{caption}
\usepackage{blindtext}
\usepackage{xcolor}
\usepackage[utf8]{inputenc}
\usepackage[T1]{fontenc} 
\usepackage{mathtools}
\usepackage{tabularx}
\usepackage{subfigure}
\usepackage{dblfloatfix}
\usepackage[justification=centering]{caption}

\usepackage{multicol}
\usepackage{lipsum}
\usepackage{mwe}
\usepackage{longtable}
\usepackage[normalem]{ulem}
\usepackage{algorithm,algpseudocode}
\usepackage{varwidth}

\begin{document}

\title{Classification of retail products: From probabilistic ranking to neural networks}
\author{Manar Mohamed Hafez, Rebeca P. Díaz Redondo, Ana Fernández-Vilas,  Héctor Olivera Pazó
\thanks{Manar Mohamed Hafez (m.mohamed@aast.edu) is with College of Computing and Information Technology, Arab Academy for Science, Technology and Maritime Transport (AASTMT)-Smart Village, P.O. Box 12676, Giza, Egypt}
\thanks{Rebeca P. Díaz Redondo (rebeca@det.uvigo.es), Ana Fernández-Vilas (avilas@det.uvigo.es) and Héctor Olivera Pazó (iclab@det.uvigo.es) are with atlanTTic, Universidade de Vigo; Vigo, 36310, Spain.}}

\maketitle

\abstract{Food retailing is now on an accelerated path to a success penetration into the digital market by new ways of value creation at all stages of the consumer decision process.  One of the most important imperatives in this path is the availability of quality data to feed all the process in digital transformation. But the quality of data is not so obvious if we consider the variety of products and suppliers in the grocery market. Within this context of digital transformation of grocery industry, \textit{Midiadia} is Spanish data provider company that works on converting data from the retailers' products into knowledge with attributes and insights from the product labels, that is, maintaining quality data in a dynamic market with a high dispersion of products. Currently, they manually categorize products (groceries) according to the information extracted directly (text processing) from the product labelling and packaging. This paper introduces a solution to automatically categorize the constantly changing product catalogue into a 3-level food taxonomy. Our proposal studies three different approaches: a score-based ranking method,  traditional machine learning algorithms, and deep neural networks. Thus, we provide four different classifiers that support a more efficient and less error-prone maintenance of groceries catalogues, the main asset of the company. Finally, we have compared the performance of these three alternatives, concluding that traditional machine learning algorithms perform better, but closely followed by the score-based approach.}


\section{Introduction}
\label{sec:introduction}
According to \cite{REINARTZ2019350} digital transformation facilitate new ways of value creation at all stages of the consumer decision process: pre-purchase (need recognition, information search, consideration or evaluation of alternatives), the purchase (choice, ordering, payment), and the post-purchase (consumption, use, engagement, service requests). This value creation is especially relevant given the very competitive nature of the retail market to gain a larger market share. Although digital transformation \cite{wessel2020unpacking} is being addressed from different fields: multi-channel solutions, user modeling, internet of things, etc. all of them rely to some extent on the availability of information on operations, supply chains and consumer and shopper behaviors. This information is the raw material for data analysis as a central driver towards digital transformation. Food retailing is now on an accelerated path to a success penetration into the digital market and manufacturers and retailers must prepare to achieve six digital imperatives.\cite{Nielsen-FMI, bahn2020descriptive}  : (1) Integrate the digital offerings with their brick-and- mortar operations; (2)  Forecasts to increase operational efficiency; (3) Optimizing omnichannel marketing and promotions; (4) Fixing inaccurate master data and (5) A single comprehensive view of customer insights; and (6) Integrating digital and in-store shelf capabilities.   

This paper focuses on the most supportive digital imperatives, that is, the availability of quality data (accurate, complete and well-maintained data) in order to feed all the visible processes in digital transformation. The quality of data is not so obvious if we consider the variety of products and suppliers in the grocery market. As a remarkable example, Walmart is the largest grocery retailer with more than 10,000 retailer units all over the world and around 3,000 suppliers and 20 million food products. Also, Walmart Marketplace gives access to third party retailers who would like to offer their products to more than 90 million unique visitors who shop on Walmart.com every month.  With this scale of locations, online users, providers, and third-party retailers, maintaining the quality of the information is an issue to address every day. Moreover, Walmart provides suppliers up-to-date sales data. Having the magnitude of Walmart or not, the challenge into a successful transition into the grocery digital market is to get the right data, process it at a high speed and obtain some valued of it.

As well as in other sectors, like financial or consumers in general, grocery industry is harnessing digital to innovate through data-drive business models. This will allow us to optimize processes and operations as replenishment and pricing and, at the same time, offer superior convenience in online grocery\cite{hafez2018comparative}. Of course, it should also be applied in brick and mortar by redesigning their strategies towards omnichannel retailing. The proliferation of these data drive models brings the emergence of data vendors also in the grocery industry. Precisely, this is where \textit{Midiadia}~\footnote{https://www.midiadia.com/} comes in. This Spanish company works on converting data from the retailers' products into knowledge with attributes and insights from the product labels. The so- called \textit{MidiadiaTECH} provides the retailers with deep knowledge about their stock and the customers with a new customized shopping experience. While \textit{Midiadia} extracts the information from the labels of the products, the categorization of the products is manually carried out. Since there are a lot of manufacturers and a lot of different sizes of shops, there is a considerable dispersion of products. Moreover, the list of products changes constantly, so it must be managed properly, taking into account the complex European legislation~\cite{eu_law} as well as the national~\cite{es_law} and regional~\cite{gl_law} legislation dealing specifically with groceries.
Therefore, the main contribution of our approach is that provides retail companies with an automatic categorization solution for new products. This classification is uniquely based on the description of each product. The methodology can be horizontally applied to different domains assuming there is a previously known taxonomy or hierarchy of categories. Thus, it would be applied to recruitment processes, travel decision support systems, investment, etc.

In this context, our objective is to provide a solution to automatically categorize the constantly changing products in the market. Being more specific, \textit{Midiadia} has defined a food taxonomy and our purpose is automatically to assign which is the right variety/category for a new product. This task must be done exclusively using the information provided by the product labeling and packaging. This would help to maintain the \textit{Midiadia} data in a more efficient and less error-prone way. With this aim, we have defined different classifiers based on three different approaches: a score-based ranking method (based on BM25), machine learning algorithms (such as K-Nearest Neighbors (KNN), fuzzy K-Nearest Neighbors (FKNN)), eXtreme Gradient Boosting (XGBoost), and deep neural networks (particularly Multilayer Perceptrons (MLP)). After comparing the results, we can conclude that if we offer only one value for the Variety, then FKNN is the best approach, but if we offer two or three options (leaving the decision between them to the experts in the company), then, the best approach is the score-based ranking method, closely followed by FKNN.

The paper is organized as follows: \autoref{sec:related} describes previous work related to classification in retailing form a customers point of view; \autoref{sec:dataset} described the dataset provided by \textit{Midiadia}; \autoref{sec:model} provides an overview of the problem to be solved. Then the four classifiers are introduced, the score-based classifier (\autoref{sec:score-based}), the nearest neighbors' classifier (\autoref{sec:NN}), an extreme gradient boosting classifier (\autoref{sec:XGBoost}), and the neural network classier (\autoref{sec:MLP}). Finally, \autoref{sec:results} shows evaluation and results and \autoref{sec:conclusions} concludes our work.

\section{Related work}
\label{sec:related}

 Automatic retail classification is an essential task for industry, since it could reduce the huge amount of human labor needed in the product chain, from distribution to inventory management. Within the traditional market field (stores and supermarkets) a considerable number of approaches in the specialized literature focus on recognizing retail products on shelves, that means, the automatic classification approaches are usually based on the product appearance. This will be very useful for retailers, who can audit the placement of products, and for costumers, who may obtain additional information about products by taking a simple picture of the shelf. These approaches usually rely on computer vision techniques \cite{baz2016context,fuchs2019towards,hafez2016effective,peng2020rp2k} and face challenging issues such as the similar appearance in terms of shape, color, texture and size of different products. In order to overcome these problems, other researches combine the information obtained from image analysis with other information based on statistical methods in order to provide a fine-grained retail product recognition and classification  \cite{baz2019statistical}. Advances in this area can help to logistic problems like shelf space planning \cite{bianchi2020retail}.

 Apart from these specific problems for image analysis, there is a well-known problem with retail products: the overwhelming amount of different products and categories in any supermarket. A typical supermarket could have more than thousand of different products \cite{goldman2019precise} and this number increases when we talk about merging catalogues from different supermarkets. Because of the popularization of e-commerce, this large number of different products are usually managed through multilevel category systems. Two relevant problems arises within this field. First, obtaining relevant information (characteristics and description) of retail products. Second, the automatic retail classification based on this data. Some approaches have focused on the first topic and try to automatically extract relevant metadata from the analysis of product images \cite{gundimeda2019automated,wang2016matching}. However, retailers usually acquire this data from suppliers and/or third parties to complete the product information on their websites. Precisely the research work introduced in this paper is being used by one of these third parties, a Spanish SME (Midiadia) that offers relevant data to retailers for their e-commerce platforms. Both, retailers and third parties, need to have automatic classification systems based on the information of the retail products to have consistent multilevel category structured data. 

 In \cite{zhong2020temporal}, this problem is tackled by applying natural language processing techniques to the text titles. One of the problems these researches face are the typical absence of grammar rules in products title or name, so they apply short text classification methods, such as character embedding, to overcome this issue. In~\cite{pobbathi2020automated}, a United States patent, a machine learning classifier helps to automatically select one or more categories for a product. The information is extracted from the metadata fields of the product description and it is used to enrich a recommendation engine. In~\cite{seth2020method}, another United States patent, a new method to automatically categorize a product in an electronic marketplace is provided. It is based on parsing the category information (provided by the retailer) to obtain a first category identifier. This information is used to search the most adequate category identifier within the available taxonomy. This methodology takes into account information of expired category identifiers. The authors in~\cite{ecommerce_auto} perform an interesting analysis of text classification and offer a solution to automated classification on e-commerce by using distributional semantics. The idea here is to create a taxonomy by first applying a modification of the Bag-of-Words model~\cite{baeza}. This approach creates feature vectors from the description of the products and combines three types of predictors: the \textit{Path-Wise Prediction Classifier}, in which each path in the taxonomy tree is considered as a class; the \textit{Node-Wise Prediction Classifier}, in which each node in the taxonomy tree is considered as a class; and the \textit{Train Depth-Wise Node Classifier}, which combines different classifiers on each level of the taxonomy tree, where each node represents a different class inside these sub-classifiers. There is also an interesting and different analysis \cite{chen2020shape} that studies how shape and packaging impact on brand status categorization, concluding that slender packages are more likely to be categorized as high-end products (high brand status) than those in short, wide packages (low brand status).

Deep learning technologies are also used in this field (retail) to enhance the information analysis processes. In fact, \cite{wei2020deep} offers a survey of the most recent works in the field of computer vision applied to automatic product recognition combined with deep learning techniques. Other interesting analysis that supplements the previous one is the work in \cite{wei2019deep}. Here the authors focus on fine-grained image analysis (FGIA) combined with deep learning to automatically assign subordinate categories in specific application fields. Although there explicitly mention species of birds or models of cars, the bases might also be applied to retail commerce as well. In the paper, the FGIA techniques are organized into three main categories: recognition, retrieval, and generation. When dealing with natural language, instead of images, there are also research works that apply deep learning to deal with automatic analysis of this kind of data. For instance in \cite{dashtipour2020hybrid} the authors face the automatic analysis of natural language to infer polarity and sentiment information from reviews in social media. Their approach is based on vector machine, logistic regression, and advanced deep neural network models. It explores dependency-based rules to extract multilingual concepts from a mixture of sentences in two languages (Persian and English) to detect subjectivity and sentiment in sentences.

Other approaches take into account the customers for gathering relevant information about products. For instance, the authors of~\cite{customer_preferences} study the influence that product categorization and types of online stores have on customers, mainly using hypothesis testing. The research follows three steps: the first one is a test to see if the online shopping preferences of the customers vary according to the type of product, without taking into account the kind of online retail store; the second step is a test on the interaction effects of both the online store and the types of products; and the last step is a hypothesis testing, which results in the identification of the attributes considered relevant by the customers when they purchase retailing products.  Along this line, the proposal in~\cite{customer_behaviour} analyses the prediction of customer behaviour in the purchase process with machine learning techniques used in real-world digital signage viewership data. The data were obtained from a camera that recorded a clothing store and additional features that were added manually. In this case, the customers can be considered in buying groups --for example, a family that is going to purchase a product. The goal was to classify the customers in six categories: (i) \textit{Initiator}, the person from the buying group who recognizes the need and finds the product that the group needs, (ii) \textit{Influencer}, the person whose opinion has a relevant effect on the purchase decision, (iii) \textit{User}, the final customer who will use the product or service, (iv) \textit{Decider}, the one who makes the final decision about whether or not to purchase a product, (v) \textit{Purchaser}, the person who pays for the purchased product and also determines the terms of the purchase, and (vi), \textit{Passive influencer}, a member of the buying group but is not involved in the purchase.  In \cite{holy2017clustering}, the approach is exclusively customer-behaviour based, providing a new clustering mechanism able to organize retail products into different categories.

In the specialized literature, there are different approaches for automatic categorization of retail products. Computer vision solutions are mainly oriented to traditional supermarkets and have to face important accuracy problems because of the huge amount of different products and the high similarity in the packaging of products within the same categories. In the E-commerce field, there are different needs, but automatic classification based on product characteristics is a relevant one. The number of different products is overwhelming, especially if we are working on combined catalogues, i.e., catalogues from different supermarkets and retailers, as it is our case. Analysing the product title or name is not enough and taking into account customer behaviour is not always available. Thus, to the best of our knowledge, there are not other studies that have faced the automatic classification problem of combined catalogues, without data from customers and only based on the product description (name, trade mark and ingredients). Our proposal fills in an interesting gap for third parties specialized on enriching the retailers catalogues with smart data.

\section{Dataset}
\label{sec:dataset}

The dataset used in this work is provided by \textit{Midiadia} which consist of a collection of $31,193$ products. The list of attributes considered in this study are shown in Table~\ref{tab:data_fields}. The {\it European Article Number} (EAN) is a standard describing a barcode symbology and numbering system used in global trade to identify a specific retail product type, in a specific packaging configuration, from a specific manufacturer.  {\it Category},  {\it Subcategory}, and  {\it Variety} represent the 3-levels (hierarchy) the company considers to organizing the catalog into a taxonomy. {\it Brand} is an identifying name for a product manufactured by a particular company. Finally, the other terms are coherent to the EU regulation, as follows. Just to clarify the meaning of these terms, a sample extract\footnote{Although product attributes are in Spanish, in the following, the translated version of fields and values are used to improve readability.} of the dataset is shown in Table~\ref{tab:real_data}:
\begin{itemize}

\item ‘Customary name’ means a name which is accepted as the name of the food by consumers in the Member State in which that food is sold, without that name needing further explanation.

\item 'Legal name' means the name of food prescribed in the Union provisions applicable to it or, in the absence of such Union provisions, the name provided for in the laws, regulations and administrative provisions applicable in the Member State in which the food is sold to the final consumer or to mass caterers.

\item An 'Ingredient’ means any substance or product, including flavorings, food additives and food enzymes, and any constituent of a compound ingredient, used in the manufacture or preparation of food and still present in the finished product, even if in an altered form; residues shall not be considered as ‘ingredients’. Moreover, the EU regulation on the provision of food information to consumers.

\item 'List of ingredients' (suitable heading which includes the word ‘ingredients’)  shall include all the ingredients of the food, in descending order of weight, as recorded at the time of their use in the manufacture of the food. 
\end{itemize}

\begin{table}[htp]
    \centering
    \caption{Product attributes in the  dataset}
    \begin{tabular}{p{45pt} p{25pt} p{130pt}}
    \hline
         \textbf{Field} &\textbf{levels}& \textbf{Description} \\
         \hline
         \textit{EAN} &  & Unique Product Number \\
         \textit{Category} & 16 & 1st Level Category  \\
         \textit{Subcategory} & 62 & 2nd Level Category \\
         \textit{Variety} & 159 & 3rd Level Category\\
         \textit{Brand} & 3,015 & Product Brand \\
         \textit{Name}& 11,139 & Product Customary Name \\
         \textit{Legal Name}  & 8,442 & Official product denomination regarding the European Union provisions \\
		    \textit{Ingredients}&  & List of Ingredients in the product \\\hline
    \end{tabular}
    \label{tab:data_fields}
\end{table}

The dataset also includes other attributes, such as packaging size, healthy claims, calories, etc. However, these attributes are aimed to recommend alternative or replacement products in a specific variety, but the problem we are trying to solve is a different one: when a new product is incorporated into the company food catalog, all the information in the product labeling  and packaging is automatically extracted but the decision of which Variety is the one that corresponds to the new product is currently manually done. Thus, we need to automatically decide to which Variety a new product belongs, so we have focused only on attributes directly related to the kind of food, discarding the others. According to this selection, rows with blank fields in all the selected columns of~\autoref{tab:data_fields} are removed. That entails that we have worked with a dataset with 20,888 products, hereinafter MDD-DS (\textit{Midiadia} DataSet).

\begin{figure}[htb] \centering
\color{black} \includegraphics[scale = 0.30]{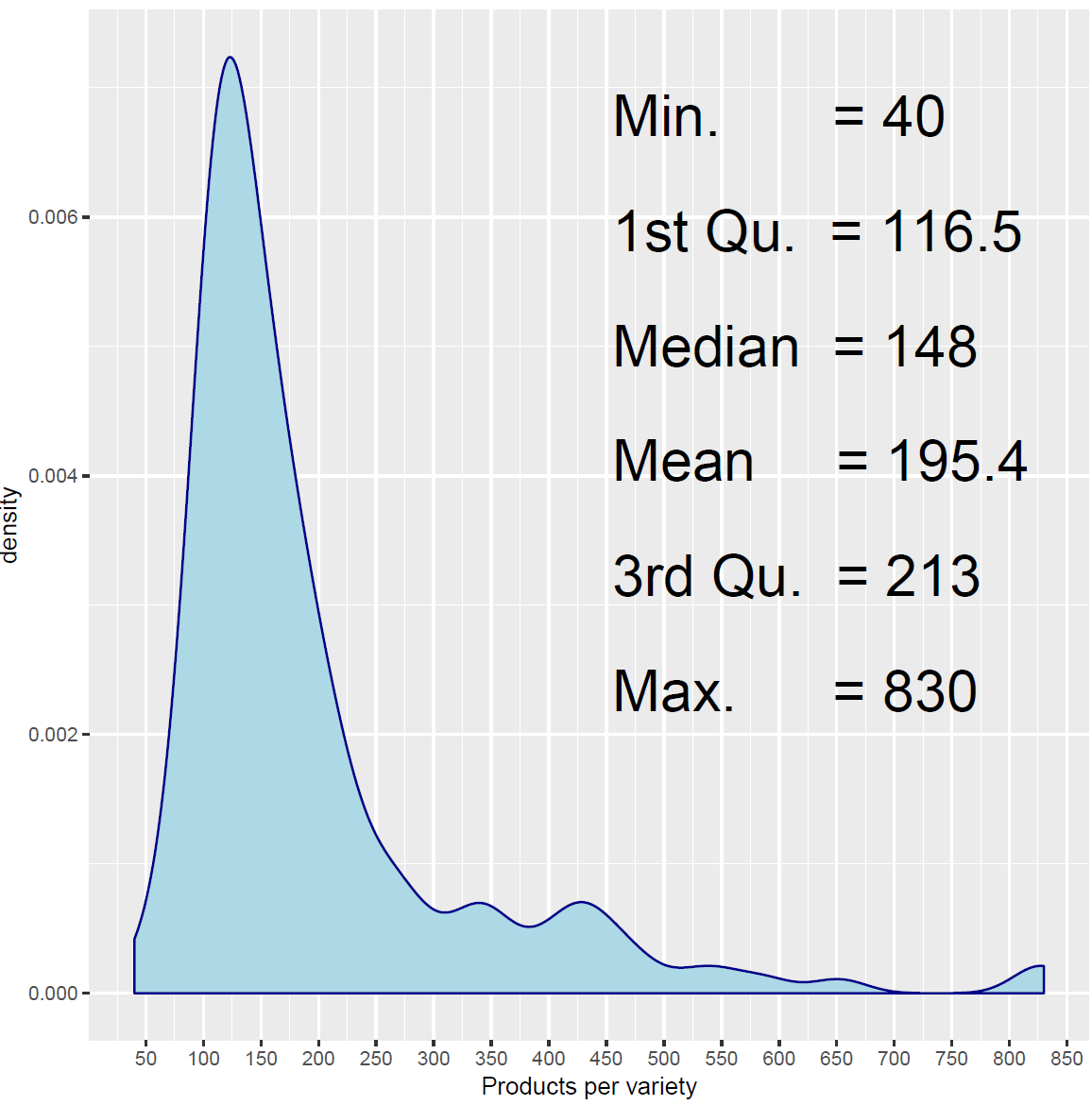} \caption{Number of products per variety: Density plot.} \label{fig:prod_vs_var_dens} \end{figure}

\autoref{tab:data_fields} also includes the number of different values that exist in the variable considered  MDD-DS. Thus, as a categorical variable, the dataset shows $159$ levels for {\it Variety}. The fact of knowing in which upper level of categorization is each variety can be helpful for classification purposes. \textit{Brand} can bring useful information apart from the {\it Name}  and {\it Legal Name} since they are defining the kind of product. Regarding {\it Ingredients}, it is a list of ingredients that require  further processing in order to support the classification.~\autoref{fig:prod_vs_var_dens} shows the distribution of products per variety as well as the statistical parameters of this distribution. The average number of products per variety is around $200$, being the 1st quartile $116.5$, so that $75$\% of the varieties have a number of assigned products above $213$ products.

\begin{table}[htb] \centering
\caption{Extract of the MDD-DS}
\begin{tabular}{p{59pt}|p{55pt}p{65pt}p{8pt}p{100pt}} \hline
\multicolumn{1}{c|}{Produt N.}&\textbf{1} &\textbf{2} & $\dots$ &\textbf{31,193}\\ \hline
\textbf{\textbf{EAN}} &62 &1292&$\dots$ &84654\\
\textbf{Category} & Fresh& Beverages&$\dots$& Snacks and nuts\\ \hline
\textbf{Subcategory}& Greens and Vegetables  & Strong Alcoholic Beverages &$\dots$ &Nuts\\\hline
\textbf{Variety}&Greens and Vegetables & Ron &$\dots$ & Seeds\\ \hline
\textbf{Mark} &Generics& Pujol &$\dots$ &Facundo\\ \hline
\textbf{Name}&Raw leek&  Golden Ron & $\dots$ & Gian Seeds\\ \hline
\textbf{Ingredients}&Leek& Golden Ron 40\% vol. alc.& $\dots$& Sunflower seeds and salt (4\%)\\ \hline
\textbf{Legal Name}&Leek& Ron& $\dots$& Roasted and salty giant sunflower seeds\\ \hline
\end{tabular}  
\label{tab:real_data} 
\end{table}

\section{Proposed Model}
\label{sec:model} 
\begin{figure}[h] \centering
\includegraphics[width=.7\textwidth]{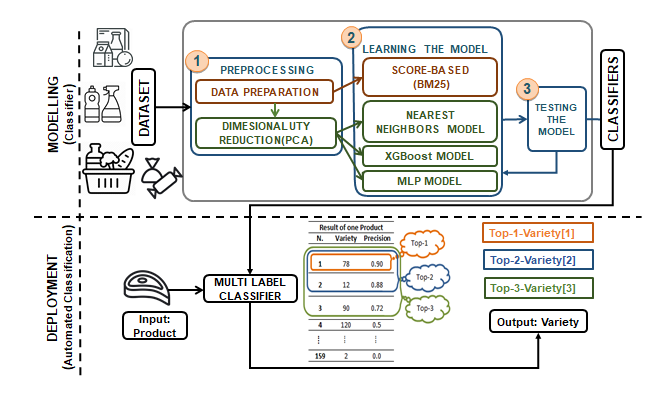} 
\caption{\label{fig:model}Description of the proposed model definition and evaluation.}  
\end{figure}

In order to automatize the time-consuming classification process, this paper proposed a multi-label classification methodology to map inputs (new products) to a maximum of 3 labels from all the varieties considered \footnote{The number of varieties is a restriction of the dataset.}. According to the scheme in~\autoref{fig:model},  the deployment of our solution (bottom part of the figure) takes  {\it Name}, {\it Legal Name}, and {\it Ingredients} of a new product ($NP$) as input observational variables. Then, a multi-label classifier predicts the varieties' score of $NP$, that is the classification score for $NP$ to be a member of each variety (class) $C_{1} \dots C_{n}$ in the taxonomy. The output is defined as the three varieties $C_{1}, C_{2}, C_{3}$  with higher scores. From these three varieties the sets $Top\_1 =  C_1$;  $Top\_2 =  \{C_1, C_2\}$ and $Top\_3 =  \{C_1, C_2, C_3\}$ are defined. This paper describes the methodology and experiments to obtain the classifiers by using the MDD-DS (described in \autoref{sec:dataset}). The modeling methodology consists of 3 main steps: (1) preprocessing the data and potentially reducing their dimensionality; (2) learning the classifier model: three different classifier models, such as score-based model, nearest-neighbors models ((KNN: K-Nearest Neighbors and FKNN: Fuzzy K-Nearest Neighbors) and deep learning (MLP or Multi-Layer Perceptron Network); (3) testing the classifier models in order to compare them and estimate the quality measures of our approach. In step (1), and in order to apply machine learning and deep learning approaches, we include a new sub-step for dimensionality reduction based on PCA (Principal Component Analysis).

The main idea of this paper is to select the best classification algorithm for retail products among a wide range of possibilities that goes from probabilistic ranking approaches to deep neural network ones. With this aim, we have selected different techniques within each approach: a score-based model in probabilistic ranking based on the BM25 algorithm; K-Nearest Neighbors (KNN), Fuzzy K-Nearest Neighbors (FKNN, a fuzzy version of the KNN classification algorithm), and eXtreme Gradient Boosting (XGBoost) in machine learning; and Multi-Layer Perceptron in deep neural networks. This selection was made taking into account the advantages of each algorithm. Consequently, BM25 was selected because of its efficiency, since its performance is well known in different ad-hoc retrieval tasks, especially those designed by TREC~\cite{okapi_exp} (see \autoref{sec:score-based} for further details). Within the machine learning field, we have selected KNN and FKNN because both of them need to perform predictions in order to learn new knowledge. Therefore, new products can be added because they will not affect the accuracy of the algorithms~\cite{abu2019effects}. In FNNN, we have proposed to use the fuzzy \textit{memberships} values of the samples to adjust the contribution values, since this algorithm delays the decision to assign a sample to a certain class across \textit{memberships}~\cite{keller1985fuzzy, derrac2014fuzzy} (see \autoref{sec:NN} for further details), and XGBoost for predictive modeling as a sturdy and efficient machine learning method for prediction. XGBoost is a boosting algorithm that belongs to supervised learning, which is an ensemble algorithm based on gradient boosted trees. It integrates the predictions of "weak" classifiers to achieve a "strong" classifier (tree model) through a serial training process. It's additionally a comparatively new technique; however, it's achieved wonderful results in several classification tasks (see \autoref{sec:XGBoost} for further details). Finally, within the deep learning area, we have selected the MLP algorithm. This is a simple but efficient algorithm that represents a broad family and that was used because of its ability to adaptive learn to perform tasks based on the data provided for the initial training or experience. Besides, it is based on a defining a decision function that it is directly obtained through training~\cite{wankhede2014analytical} (see \autoref{sec:MLP} for further details).

\subsection{Preprocessing}

As previously mentioned, MDD-DS consists of 20,888 products, which is the result of cleaning the original dataset (31,193 products) before applying the classification algorithm. This first cleaning procedure consisted of removing all the products with no name, no ingredients, and no legal name. After that, MDD-DS is pre-processed by extracting all the meaningful words for the attributes 'Name', 'Legal Name'  and 'Ingredients'. For each product $p$, the three attributes are merged into a single text which aims to describe the product $des(p)$. This description ($des(p)$) is obtained after a cleansing process described\cite{al2013lemmatizing} step by step in Algorithm~\ref{alg:preprocess}: (i) parenthesis are transformed into blank spaces, (ii) numbers, stop-words, punctuation and extra spaces are removed (iii) all letters are converted to lowercase, and (iv) repeated strings are removed. 

\begin{algorithm}[h]

\caption{Preprocessing pseudocode \label{alg:preprocess}}
{\small
	\begin{algorithmic}[1]
\Procedure {Preprocess}{MDD-DS}
\State cleaning($MDD\_DS$)\;
\State  $product\_words[] \gets$ new\_list($m$) \;
\State  $all\_products\_words \gets$ new\_vector (0) \;
\For{i  $\gets$ 1 : m} 
\State $p \gets$ MDD-DS$[i,\ ]$\;
\State $des(p)$ $\gets$ concatenate(\;
\State $p$[Name], $p$[Legal Name], $p$[Ingredients])\;
\State $des(p)$  $\gets$ transform\_brackets\_into\_space($des(p)$)\;
\State $des(p)$  $\gets$ transform\_into\_lowercase($des(p)$)\;
\State $des(p)$  $\gets$ remove\_numbers($des(p)$)\;
\State $des(p)$  $\gets$ remove\_stopwords($des(p)$)\;
\State $des(p)$  $\gets$ remove\_punctuation($des(p)$)\;
\State $des(p)$  $\gets$ remove\_extra\_spaces($des(p)$)\;
\State $des(p)$  $\gets$ split\_by\_spaces($des(p)$)\;
\State $des(p)$  $\gets$ remove\_duplicates($des(p)$)\;
\State $des(p)$  $\gets$ remove\_empty\_words($des(p)$)\;
\State $product\_words[i]$  $\gets$ $des(p)$\;
\State $all\_products\_words$ \;
\State $\gets$ $all\_products\_words$  $\cup$ $des(p)$\;
\EndFor
\EndProcedure
\end{algorithmic}}
\end{algorithm}

Then words are split and a vector of words $product\_words(p)$  is generated, an example is shown in~\autoref{tab:prod_ing}. Additionally, we also create the $all\_products\_words$ (for all the products in MDD-DS). This is another vector of words that contains as many elements as different words in the products' vectors (from the fields Name, Legal Name, and Ingredients). After preprocessing, $all\_products\_words$ contains a total of 11,359  unique words, an example is shown in~\autoref{tab:prod_ing_1}.

\begin{table}[h] 
\begin{center}
\caption{Examples of $product\_words$ for every $p$}
\begin{tabular}{ll} \hline
	product id & product-vector \\
	\hline
	1 & ['oil', 'olive', 'virgin', ...] \\
		$\dots$ & $\dots$ \\
	218 & ['pepper', 'sweet', 'glass', ...]\\
	$\dots$ & $\dots$ \\
	$20,888$ & ['biochips', 'cheese',  'spelt', 'oil', ...]\\\hline
	
\end{tabular}  
\label{tab:prod_ing}
\end{center}
 \end{table} 

\begin{table}[h] \centering
\caption{Extract from $all\_products\_words$. \label{tab:prod_ing_1}  }
\begin{tabular}{|p{\columnwidth}|} 
	\hline
['oil', 'olive', 'virgin', 'extra', 'arbequina', 'extraction', 'cold', 'filter', 'quality', 'high', 'origin', 'gluten', 'fondant', 'picual', 'spanish',  'refined',  'e420ii', 'e1520', 'liquid egg', 'gluten',  'e501i', 'e281', 'powder', 'bicarbonate',  'fructose',  'e218', , 'icing',  'e492',  'fondant', 'aluminum', 'e155', $\dots$]\\
\hline
\end{tabular}  
\end{table}

With the information provided by the $all\_products\_words$, we obtain the product matrix $X[m,n]$, which is mathematically defined as follows. Let $\vec{w}$ be the $n$-dimensional vector obtained from $all\_products\_words$ such that $\vec{w} = (w_1,\dots,w_n)$ and $\forall\;k\in [1,\dots,n],\;w_k$ is a string $\in$ $all\_products\_words$ and $N=\dim(all\_products\_words)$ is the total number of different meaningful words in the dataset (after cleaning).

\begin{itemize}  
	\item The product matrix $X[m,n]$ is a $MxN$ matrix where each column represents a word $w_i \in \vec{w}$ and each row represents a product $p$ of the dataset so that $M$ is the total number of products 

	\item Let $\vec{p}$ be the $product\_words[]$  list such that, each element contains a vector $\vec{p}_i$ hence $\vec{p} = (\vec{p}_1, \dots, \vec{p}_{m})$. Likewise, $\vec{p_i} = (p_{i}[1], \dots, p_{i}[l])$ where $l=\dim(product\_words[i])$, length of the vector in the $i$ element of the list $\vec{p}$. Note that $\forall\;k\in [1,\dots,l],\;p_i[k]$ is a string. 
	
	 \end{itemize} 

Therefore: 

\begin{equation} X[i,j] = \Bigg\{ \begin{tabular}{l} 1, if $\vec{w}[j]$ $\in$ $\vec{p}[i]$\\ 0, if $\vec{w}[j]$ $\notin$ $\vec{p}[i]$\\ \end{tabular} \end{equation}
	
where 

\begin{equation}\vec{w}[j]\;\in\;\vec{p}[i]\;\iff\;\exists\;k\;/\;\vec{p}_{i}[k]\;=\;\vec{w}[j]\end{equation}

This means $X[i,j]$ is $1$ if the product $i$ has the word $j$ in its \textit{product\_words[i]}; otherwise, the value will be $0$. 
For evaluation purposes, we include the variety of the product in an additional column, as shown in~\autoref{tab:prod_mat}. 

\begin{table}[h] \centering
\caption{Example of a \textit{product matrix}.  \label{tab:prod_mat} }
{\small
\begin{tabular}{lllllp{2cm}}\hline 
	$p$ id &\textbf{'oil'} &\textbf{'olive'} & \textbf{'virgin'} & $\cdots$ &\textbf{Variety}\\ \hline
\textbf{1}& 1 & 1 & 1 & $\cdots$ & 'Virgin olive oil and extra virgin olive oil' \\ \hline
$\vdots$ &$\vdots$ & $\vdots$ & $\vdots$ & $\dots$ & $\vdots$\\ \hline
\textbf{20,888} & 1 & 0 & 0 & $\dots$ & 'Salads' \\ \hline
\end{tabular} }
\end{table}

There are some interesting techniques within this field, such as Principal Component Analysis (PCA)~\cite{wang2020principal}, Linear Discriminant Analysis (LDA)~\cite{balakrishnama1998linear}, and Autoencoders~\cite{kunang2018automatic}. PCA and LDA are both linear transformation techniques used for dimensionality reduction and visualization, but while PCA is an unsupervised technique, LDA is a supervised one. PCA searches for attributes with the most variation and ignores class labels. In contrast, LDA maximizes the separation of known categories.  An autoencoder is a neural network, with as many output units as input units and at least one hidden layer, that is used to learn a representation (encoding) for a data set by training the network to avoid or ignore signal noise. This provides a non-linear transformation that reduces the dimensionality of the data. The authors of~\cite{wang2016auto} evaluate the performance of this technique by comparing an autoencoder with other models that can perform a similar reduction task, such as PCA, LDA, etc. The results showed that PCA and  LDA are relatively more stable than autoencoder (low stability implies that different results might be obtained when replicating the experiment). To sum up, we opted for an stable reduction technique and, more precisely, an unsupervised one, that is PCA. Both stability and the absence of a training step before applying reduction eases the  deployment of automatic classification in a real retailing scenario, where the product catalogue is intrinsically dynamic.

PCA uses an orthogonal transformation to convert a set of observations of possibly correlated variables into a set of linearly uncorrelated variables. In addition, PCA collects all the required features for MDD-DS, which removes correlated features~\cite{jung2003principal}, improves algorithm performance~\cite{wang2020principal}, reduces over-fitting~\cite{reddy2020analysis}, and improves visualization~\cite{ reddy2020analysis}. To achieve the research goal, our study compares the results obtained with different values for PCA (400, 500, 600,700, and 800) to select the best option (in \autoref{sec:NN} and \autoref{sec:MLP}).

\section{Score-based Model}
\label{sec:score-based}
Our first approach is based on the bag-of-words BM25 model~\cite{bm25_trec5} (BM for Best Matching), widely used in information retrieval to rank documents based on a query of words. The BM25 equation considers the frequency of occurrence of the words in the documents, smoothing out the weighting in favour of how distinctive they are: 

\begin{equation} \sum_{w \in Q \cap D} \frac{(k + 1) c(w,D)}{c(w,D) + k(1 - b + b \frac{|D|}{avdl})} \cdot \log \frac{N - df(w) + 0.5}{df(w) + 0.5},\label{eq:bm25} 
\end{equation} 

where \textit{w} is the word analyzed, \textit{c(w, D)} is the number of times the word \textit{w} appears in the document \textit{D}, \textit{Q} is the query, \textit{N} is the number of documents, \textit{D} is the document analyzed, \textit{df(w)} is the number of documents that contain the word \textit{w}, \textit{k} is a constant with the value 1.2 (typical value obtained from TREC experimentation~\cite{okapi_exp}), $|D|$ is the total number of words in the document \textit{D}, \textit{b} is a constant with the value 0.75 (typical value obtained from TREC experimentation~\cite{okapi_exp}), and \textit{avdl} is the average number of words per document. 

In order to apply this probabilistic model to the classification of retail products into varieties, after obtaining the \textit{product matrix}, we generate a \textit{variety matrix}. This matrix will have as many columns as different words exist in the dataset --i.e., the same number of columns as the \textit{product matrix}. The number of rows will be the same as the number of different  varieties obtained from the dataset.

\begin{itemize} \item Let $Y$ be the \textit{variety matrix} ($y \, x \, n$). \item Let $\vec{z}$ be the $y$-dimensional vector of integers where each one represents a variety. Thus, $\vec{z} = (z_1,\dots,z_y)$. \item Let $\vec{v}$ be the $m$-dimensional vector that contains an integer that represents each product variety. Thus, $\vec{v} = (v_1,\dots,v_y)$. \end{itemize} 

Therefore: 

\begin{equation} Y[i,j] = \sum_{\vec{v}[k] = \vec{z}[i]} X[k,j] \label{eq:var_mat}\end{equation} 

In \autoref{eq:var_mat}, the value of $Y[i,j]$ will be the number of times the word $j$ appears in the \textit{variety} $i$ (i.e., the set of all the products belonging to it and represented by the index $k$), as shown in~\autoref{tab:var_mat}. 

\begin{table}[h] \centering
\caption{Example of a \textit{variety matrix}.\label{tab:var_mat}}
\begin{tabular}{lllll} \hline \

	 &\textbf{'oil'} &\textbf{'olive'} &$\cdots$ &\textbf{'aromatic'	}\\ \hline
\textbf{Almonds / hazelnuts}& 0 &0 &$\dots$ &0\\ \hline
$\vdots$ &$\vdots$ &$\vdots$ &$\cdots$ &$\vdots$\\ \hline
\textbf{Virgin and Extra virgin Olive Oil} & 233 & 224 & $\dots$ & 0\\ \hline
$\vdots$ & $\vdots$ & $\vdots$ & $\cdots$ & $\vdots$\\ \hline
\textbf{Juices (100\%)} & 0 & 0 & $\cdots$ & 0\\ \hline
\end{tabular}   
\end{table} 

To obtain the variety \textit{V} of a product, we apply the modification of the BM25 model depicted in (\autoref{eq:bm25_mod}). This equation gives a realistic smoothing when the term frequency is high and rewards low-frequency words even more than the one in (\autoref{eq:bm25}). We consider two modifications with regard to the original BM25 equation. First, one unit is added to both the nominator and the denominator. The reason for the addition in the denominator is that the term of the query may be missing in the analyzed document within the corresponding iteration of the addition, so $vf(w)$ can be equal to zero. Adding a unit instead of 0.5 makes more sense in our scenario because it is more realistic to add integers (i.e., entire words) than a rational number; the lower bound is also equal to the unit. At the same time, the unit added to the numerator balances out the fraction. The second modification is that the factor $c(w,V)$ is a constant. This is due to the fact that the word duplicates contained in the query are removed and the terms are unique. Thus, $c(w,V)$ will always be equal to 1.

\begin{equation}  \sum_{w \in P \cap V} \frac{(k + 1) c(w,V)}{c(w,V) + k(1 - b + b \frac{|v|}{avvl})} \log (\frac{N + 1}{vf(w) + 1}) \label{eq:bm25_mod}\end{equation} 

Once the modified BM25 equation is repeated for all the varieties, the variety was chosen as a result for a product is the one ranked the highest. 

\section{Nearest-Neighbors Model}
\label{sec:NN}

The nearest neighbor (NN) \cite{rastin2020generalized} rule is a nonparametric method for classification based on instances. Taking into account the problem in~\autoref{fig:model},  a product in the MDD-DS $p$ follows the  definition $\{p_1, p_2, .., p_n, v\}$ where  $n$ is the number of the aforementioned product attributes  and $v$ is it's assigned product variety.

The KNN approach ~\cite{abu2019effects}  classifies an unlabelled product on the majority of similar sample products  among the k-nearest neighbors in MDD-DS that are the closest to the unlabelled product.  The distances between the unlabelled product and each of the training product samples is determined by a specific distance measure. Therefore, the variety of new product $\vec{p}$ is assigned to the most common variety  among its closest $k$ training samples according to the attributes considered in the product vectors.  More formally, let $\vec{unlabel-p}$  an unlabelled product, the decision rule predicts a variety $\hat{v}$ for the $\vec{unlabel-p}$ according to the variety $v$ of the majority of its $k$ nearest neighbors; in the case of a tie, $\hat{v}$ is given by the closest nearest neighbor that belongs to one of the tied varieties.

One of the problems for KNN is to give equivalent importance to each of the samples to determine the class membership, neglecting the typicality among them. Alternatively, some challenges arise for classification in high dimensionality dataset where it is complex to discriminate between classes. This problem is due to the fact that the number of training samples and the increase in dimensionality grow overwhelmingly. The aforementioned drawbacks have been addressed in the literature by Fuzzy Sets Theory to allow imprecise knowledge to be represented and fuzzy measures to be introduced.

Consequently, FKNN is a fuzzy version of the KNN classification algorithm~\cite{keller1985fuzzy}. Fuzzy classifiers delay the decision to assign a sample to a certain class by using \textit{memberships}. In terms of our problem in~\autoref{fig:model}, FKNN assign to an $\vec{unlabel\_p}$  a variety's fuzzy membership (not just true or false) by taking into account the \textit{distance} from $\vec{unlabel\_p}$ to its $k$ nearest neighbors and those neighbors' memberships for this variety. More formally:

\begin{itemize} 
\item Variety membership is defined for each product $\vec{p}$ in MDD-DS with a value in $[0, 1]$ for each variety $v$. Although the details can be found in \cite{keller1985fuzzy}, the main idea is assigning to each product $\vec{p}$ a membership value for a variety $v$ not just according to the assigned variety (true value) but also according to the assigned variety for its $k$ nearest neighbors.
\item Besides given an unlabeled product $\vec{unlabel\_p}$, each neighboring product $\vec{p}=\{p_1, p_2, .., p_n, v\}$ votes for every variety $\{v_1, v_2, \dots, v_m\}$ by using its variety memberships (not just for its assigned one $v$). Again  the details can be found in \cite{keller1985fuzzy} but votes are weighted according to the inverse of the distance to  $\vec{unlabel\_p}$ and  added. The estimated variety  $\hat{v}$ for $\vec{unlabel\_p}$ is the variety with the greatest combined vote.
\end{itemize}

The main benefit of using the FKNN model may not be decreasing the error rate but, more importantly, the model provides a degree of certainty that can be used with a "refuse to decide" option. Therefore, objects with overlapping classes can be detected and processed separately

Since KNN and FKNN are computationally expensive algorithms, we apply Principal Component Analysis (PCA) to MDD-DS to reduce its  dimensionality. In our experiment, the original set of product attributes are transformed into a smaller set (the so-called principal components) so that the reduced dataset MDD-DS-reduced  still contains most of the information in MDD-DS. More specifically,  we evaluate the performance of KNN and FKNN applying PCA and selecting  400, 500, 600, 700 and 800 principal components. As a result, the \textit{product matrix} is replaced by a PCA-reduced matrix. Given the variance retention for the following number of eigenvectors $ (400, 0.796452);$ $(500, 0.825814);$ $(600, 0.84735);$ $(700,0.865117);$ $(800, 0.879153)$, the experiments in this paper have been carried out over a reduced dataset of $600$ principal components, referred to as MDD-DS-PCA-600. 

\subsection{Optimal $k$ and  \textit{Distance} for KNN and FKNN}
The key to success in KNN and FKNN are finding the optimal value of $k$ and  the distance function. Usually, the $k$ parameter is chosen empirically depending on each problem, different numbers of nearest neighbors are tested, and the parameter with the best  performance (accuracy) is chosen to define the classifier. Choosing the optimal $k$ is almost impossible for a variety of problems since the performance of a KNN classifier varies significantly when changing $k$ and the change of the \textit{distance} metric used~\cite{hassanat2014solving}.
\begin{figure}[htb]
    \includegraphics[width=0.4\textwidth]{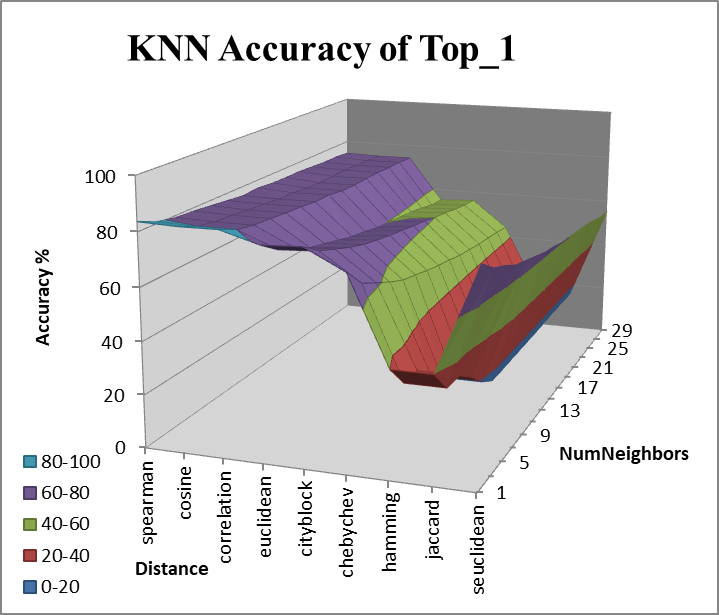}\hfill
    \includegraphics[width=0.4\textwidth]{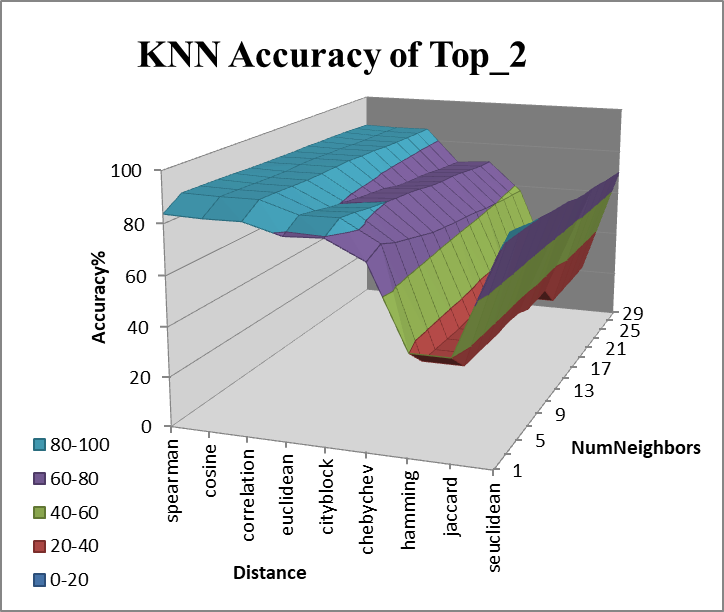}\hfill
    \includegraphics[width=0.4\textwidth]{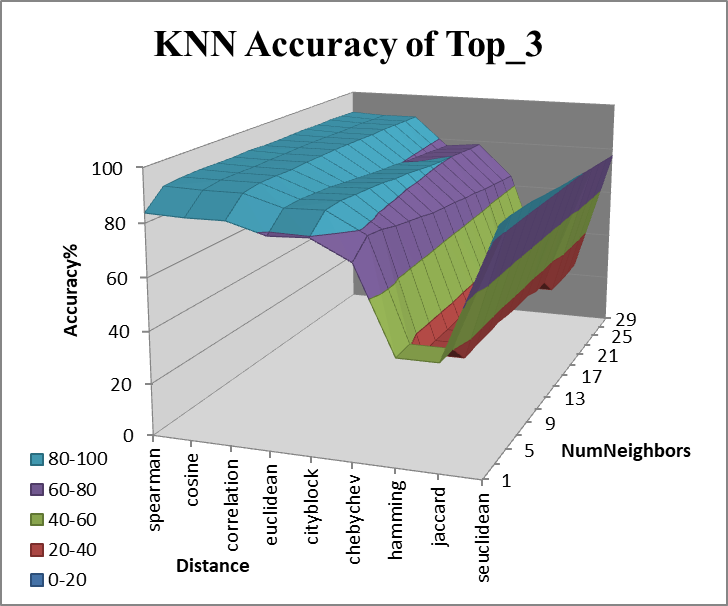}
    \includegraphics[width=0.4\textwidth]{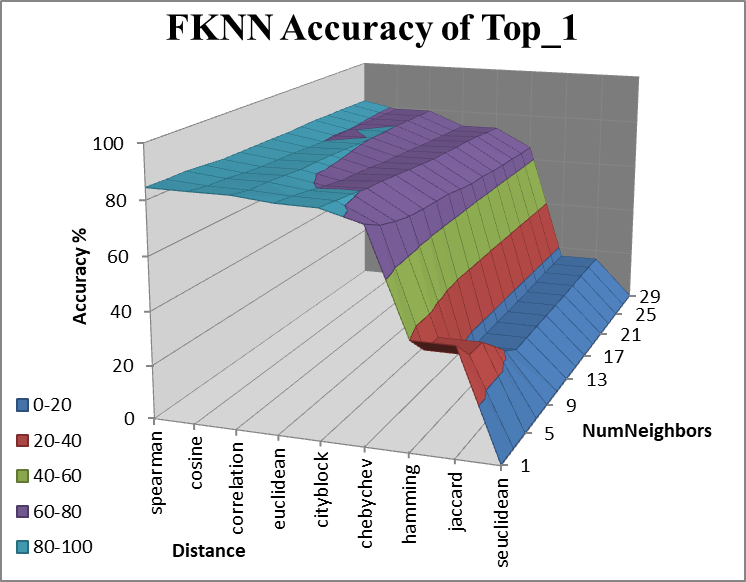}\hfill
    \includegraphics[width=0.4\textwidth]{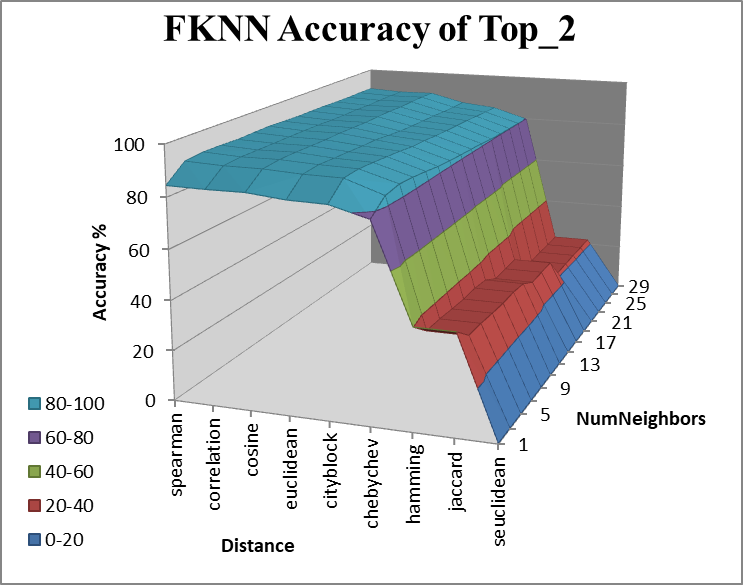}\hfill
    \includegraphics[width=0.4\textwidth]{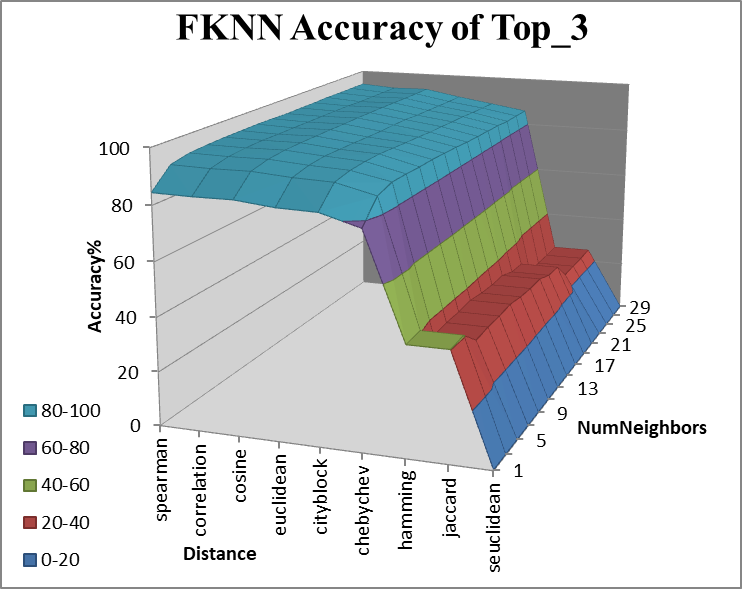}
    \caption{Accuracy of KNN and FKNN for the three approaches: $Top\_1$, $Top\_2$ and $Top\_3$}
\label{fig:knn:Fknn_all}
\end{figure}
Therefore, the decision about the \textit{distance} measure also plays a vital role in determining the final result of the classification. Although Euclidean distance is the most widely used distance metric in NN-based classifiers, we have applied a variety of  \textit{distance} metrics in MDD-DS. We refer to as "best distance metric" to the metric which allows KNN, FKNN to classify test samples with the highest accuracy, recall, precision, and F-score. The definition of the distances considered in this paper (Spearman, Cosine, Correlation, Euclidean, Cityblock, Chebychev, Hamming, Jaccard, and Seuclidean) are described in~\autoref{tab:distance_measure}, included as an appendix.

\begin{table}[ht] \centering
\caption{Optimal Parameters of \textit{KNN, FKNN} for each approach.}
\begin{tabular}{ccc|cc}
\hline
   & \multicolumn{2}{c}{KNN} &  \multicolumn{2}{c}{FKNN}\\
	& $k$ & \textit{Accuracy} & $k$ & \textit{Accuracy} \\ \hline
    \textbf{$Top\_1$}& 1 &0.8367 & 1 & 0.8446 \\ 
    \textbf{$Top\_2$}& 3 &0.8874 & 13 & 0.9153\\ 
     \textbf{$Top\_3$}& 7 &0.9141 & 23 &0.9413\\ \hline
\end{tabular} 
\label{tab:result_Knn} 
\end{table} 

To determine the optimal value of $k$ and \textit{distance} for  Nearest-Neighbors Models KNN and FKNN, we deploy the pseudocode in Algorithm~\ref{alg:alg1}  to try different combinations in  two loops (odd $k$ values initialized from $[1\ to\ 30]$ and  the different \textit{distances} used) over the reduced dimension MDD-DS for 600 principal components. This value was selected taking into account that with 600 principal components we obtain a value around 85\% of retained variance, as a criteria to choose the appropriate number of principle components. Besides, we have perceived that the results do not significantly improve if the number of principal components increase. Since we pursue to obtain the optimal pair of $k$ and $distance$ not only for classification per se ($Top\_1$) but also for $Top\_2$ and $Top\_3$, the pseudocode obtain  the accuracy for all the combinations as it is shown in~\autoref{fig:knn:Fknn_all}.

The optimal type of the \textit{distance} is Spearman of three approaches of KNN and FKNN. \autoref{tab:result_Knn} shows that the optimal value of $k$ and accuracy result for each approach according to the~\autoref{fig:knn:Fknn_all}. 

\begin{algorithm}[htp]
\caption{Tuning parameter for KNN and FKNN \label{alg:alg1}}
{\small 
\begin{algorithmic}[1]
\Procedure {Tuning\_Parameters}{Train\_PCA600, Test\_PCA600}
\State $Dist[]$ = ['Spearman', 'Cosine', 'Correlation', 'Euclidean', 'Cityblock', 'Chebychev', 'Hamming', 'Jaccard', 'Seuclidean'] 
\State Label $C1$  \Comment{First Variety}
\State Label $C2$  \Comment{Second Variety}
\State  Label $C3$ \Comment{Third Variety}
\State i $\gets$ 1 
\For{K $\gets 1:30$ by $2$} 
     \For {$J \gets 1:9$} 
       \State $M \gets$ build\_model(Train\_PCA600, $k$, $Dist[J]$)
       \State $[C1,C2,C3] \gets$ Predict($M$ , Test\_PCA600)
   \EndFor
        \State $i \gets i+1$
\EndFor
\State $Top\_1 \gets C1$
\State $Top\_2 \gets  C1 \  \And \ C2$ 
\State $Top\_3 \gets C1 \  \And  \ C2 \ \And \  C3$ 
\State \textbf{return} ${Top\_1,\  Top\_2,\  Top\_3}$ 
\EndProcedure
\end{algorithmic}}
\end{algorithm}

\section{XGBoost Model}
\label{sec:XGBoost} 
eXtreme Gradient Boosting (a.k.a XGBoost)~\cite{thongsuwan2021convxgb,shilong2021machine} is an extremely scalable end-to-end tree boosting system, a machine learning technique for classification and regression issues. XGBoost is additionally boosting algorithm belonging to supervised learning; it uses an ensemble of $X$ classification and regression trees (CARTs), each inside node represents values for attributes test, and a leaf node with scores represents a choice ($X_{E}^i|i \in 1..X$).

In terms of our problem in~\autoref{fig:model}, and given $\vec{p}=\{p_1, p_2, .., p_i, \hat{v}\}$, XGBoost can learn using~\autoref{eq:xgboost}, which is $(p_1, p_2, .., p_i)  \rightarrow \hat{v}$.  Given a set of features $\{p_1, p_2, .., p_i\} $ for $\vec{unlabel\_p}$, XGBoost can predict based on the total of the prediction scores for every tree:
\begin{equation} 
\hat{v_i}=\sum_{x=1}^{X} s_x(p_i), s_k \in S
\label{eq:xgboost}\end{equation} 
where $p_i$ represents the $i^{th}$ training sample of MDD-DS and $v_i$ are the variety corresponding label, $s_x$ represents the score for $x^{th}$ tree, and $S$ represents the set of all $X$ scores for all CARTs. 
XGBoost adopts a similar gradient boosting as the Gradient Boosting Machine (GBM). Regularization is applied to enhance the ultimate result:
\begin{equation} 
LF(\theta)=\Sigma_{i} l(\hat{v_i},v_i)+	\Sigma_{x} \Omega(s_x)
\label{eq:xgboost1}\end{equation} 
where $LF(\theta)$ means the total objective function, $l$ represents the differentiable loss function, that measures the difference between the prediction of a variety $\hat{v_i}$ and the target label of a variety $v_i$. The $\Omega$ avoids over-fitting penalizes the complexity of the model:
\begin{equation} 
	\Omega(s)=\gamma{T}+\frac{1}{2}\lambda \sum_{j=1}^{T} w_{j}^{2}
\label{eq:xgboost2}\end{equation} 
where $\gamma$ and $\lambda$ are constants that control the degree of regularization, $T$ represents the leaves number in the tree, and $w$ means the weight of each leaf. 
XGBoost also applies a second-order approximation to extend the loss function and removes the constant term and applies two additional techniques to reduce overfitting further; the details can be found in~\cite{thongsuwan2021convxgb,shilong2021machine}.

\section{MLP Model}

\label{sec:MLP}
\begin{figure}[htp]
  \centering
  \includegraphics[width=.55\textwidth]{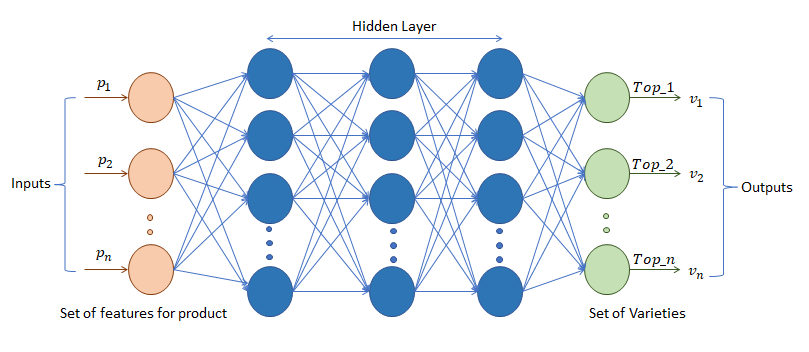}  
  \caption{MLP structure.   \label{fig:MLP:model}}
\end{figure}

Multilayer perceptron’s (MLP) \cite{feng2020multi} is a deep feed-forward neural network particularly designed for multi-class supervised learning problems. It has one or more hidden layers between the input and the output, each layer fully connected to the next. An MLP network is trained on a set of input-output pairs so that the network learns to model the dependencies between those inputs and outputs. Training involves adjusting the parameters, or weights and biases, of the model to minimize error. Back-propagation is used to make those weight and bias adjustments in relation to error and error itself can be measured in various ways.~\autoref{fig:MLP:model} shows MLP structure with scalar output.

In terms of our problem in~\autoref{fig:model}, and given $\vec{p}=\{p_1, p_2, .., p_n, v\}$, MLP learns a function $f(\{p_1, p_2, .., p_n\})  \rightarrow v$.  Given a set of features $\{p_1, p_2, .., p_n\} $ for $\vec{unlabel\_p}$, MLP  can learn a non-linear function for classification in the  varieties space $v$. For the purpose of comparison, we built-up an MLP network from 600 principal components as a reduced input set. With a Leaning Rate \footnote{Learning rate is a hyper-parameter that controls how much we are adjusting the weights of our network with respect the loss gradient.} of 0.001 which is its default value, the first experiments showed a significative increase of performance from 1 to 3 hidden layers and, above 3 hidden layers (3,4,5), the increase in performance slows down meanwhile computation time increases significantly.  

\begin{figure}[ht!]
    \includegraphics[width=0.5\textwidth]{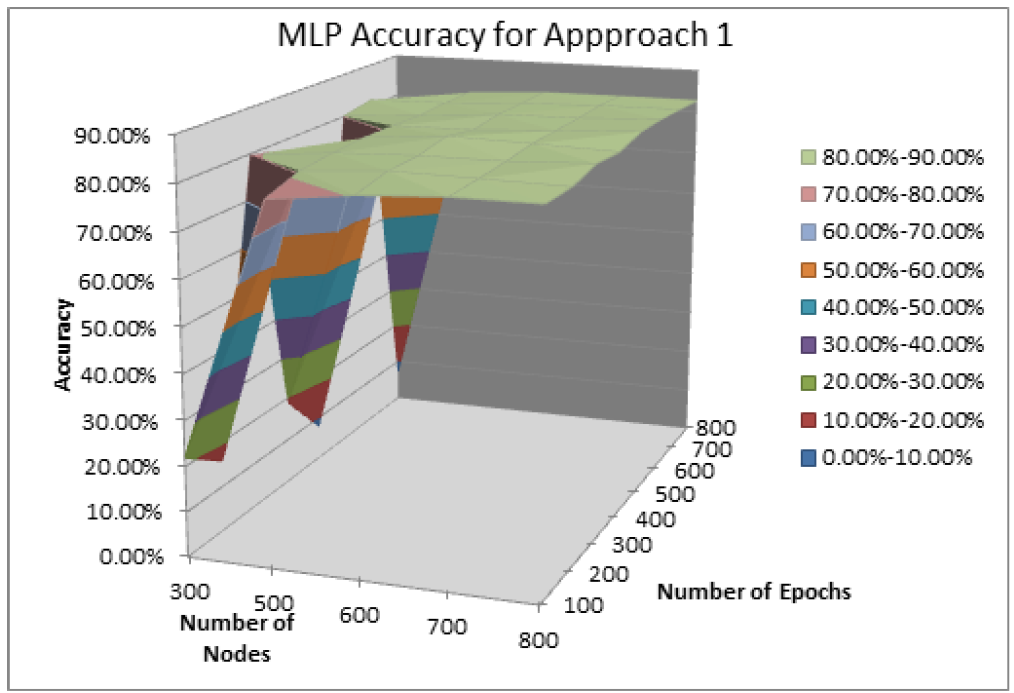}\hfill
    \includegraphics[width=0.5\textwidth]{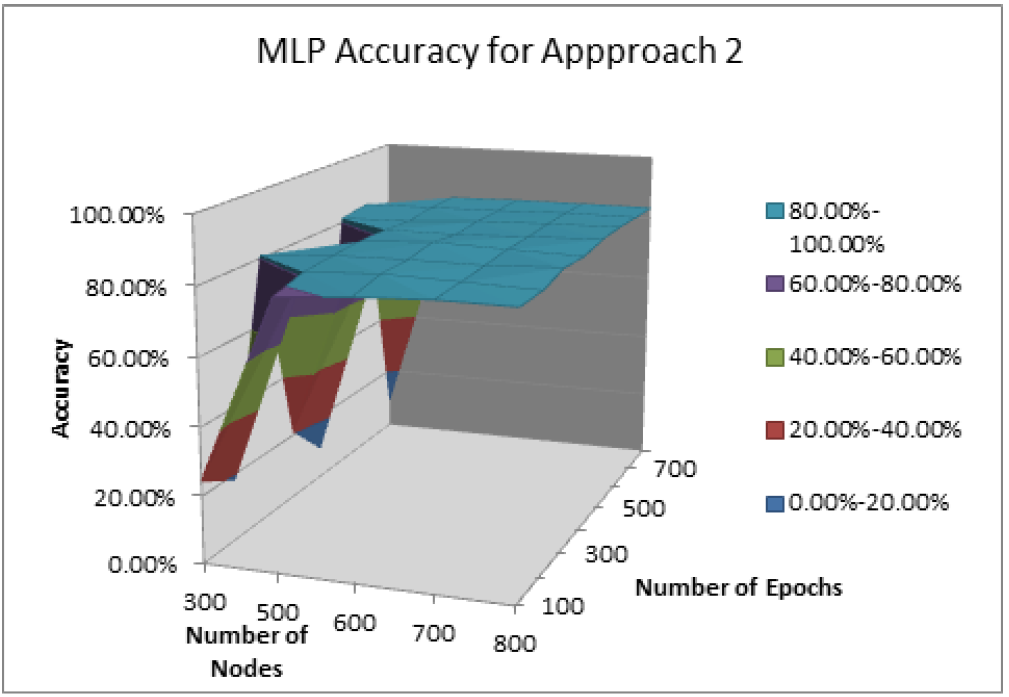}\hfill
    \includegraphics[width=0.5\textwidth]{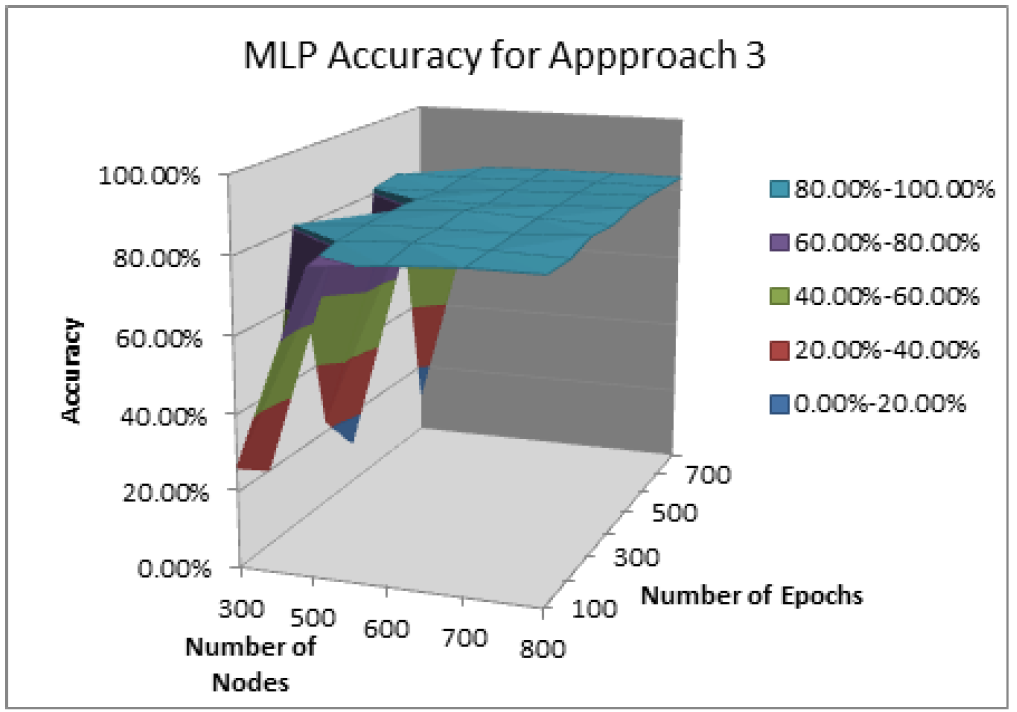}
\caption{Accuracy of MLP for the three approaches: $Top\_1$, $Top\_2$ and $Top\_3$}
\label{fig:mlp_all}
\end{figure}

In practice, as the number of hidden layers determine the ability to generalize, we selected $3$ hidden layers, upper values, apart from increasing computation time, tend to overfit which leads to reduce classification performance for  out-of-sample instances.  To obtain the optimal values of the number of nodes per layer $N$ and the times an algorithm visits the data set, training epoch $T$, we applied the pseudocode in Algorithm~\ref{alg:alg1_Mlp} over the reduced dimension MDD-DS for 600 principal components (as mentioned, the reason in~\autoref{sec:NN}). According to the curve in~\autoref{fig:mlp_all} accuracy increases when the number of training epochs significantly increase, and remarkably does not increase when applying 600, 700, and 800 and the same for the number of nodes. Therefore, the two  loops can stop at $800$;  number of nodes $[300,400,500,600,700,800]$ and training epochs  $[100, 200, 300, 400, 500, 600, 700, 800]$. The result of the evaluation shows that the optimal number of nodes is 800 and the number of training epochs is 600 for the three approaches $(Top\_1, Top\_2$ and $Top\_3)$.

\begin{algorithm}[htp]
\caption{Tuning parameter for MLP to the reduce MDD-DS with 600 principal components and with a total of $159$ classes (varieties) \label{alg:alg1_Mlp}}
{\small 
\begin{algorithmic}[1]
	\Procedure{Tuning\_Parameters}{Train\_600,\  Test\_600}
	\State $Learning\_Rate LR = 0.001$;  $Hidden\_layer HL= 3$
\State $Number\_of\_Node\ N$ ; $Training\_epochs\  T$;
\State Label $C1$  \Comment{1st Variety}\;
\State Label $C2$  \Comment{2nd Variety}\;
\State  Label $C3$ \Comment{3rd Variety}\;
\For{$N \gets 300:800$ by $100$} 
 \For{$T \gets 100:800$ by $100$}
    \State $Model \ M  \gets$ build\_model\_MLP(Train\_600, Test\_600, $HL$, $LR$ )\;
     \State $[C1,C2,C3] \gets$ Predict($M$ , Test\_600)\;
      \State $ T \gets T+1$
 \EndFor
    \State $N \gets N+1$
\EndFor
\State $Top\_1 \gets C1$ \; 
\State $Top\_2 \gets C1 \  \And \  C2$ \; 
\State $Top\_3 \gets C1 \  \And  \ C2 \ \And \  C3$ 
\State \textbf{return} ${Top\_1,\  Top\_2,\  Top\_3}$ \;
\EndProcedure
\end{algorithmic}
}
\end{algorithm}

\section{Evaluation and results}
\label{sec:results} 

For evaluation purposes, we have implemented the approaches using different tools and programming languages: (i) R (Rstudio) to pre-process the data set and then (ii) Matlab and Python to implement machine learning algorithms and deep neural networks. We have analyzed the usual parameters of each algorithm: (i) $k$ and the distance for KNN and FKNN, because these two parameters are the ones that condition the algorithm~\autoref{sec:NN}; and (ii) the number of hidden layers, the number of nodes and the number of epochs for MLP are the three most relevant parameters that condition the algorithm of neural networks ~\autoref{sec:MLP}.

In order to evaluate the different approaches, we randomly split the data from the \textit{product matrix} obtained from the dataset into a training set ($\simeq$ 90\% of the data) and a testing set ($\simeq$ 10\% of the data). The evaluation performed on some of the measure parameters, such as accuracy, precision, recall and $F1$-score~\cite{measure_parameter} as shown in \autoref{eq:ev:formulas}. These parameters are obtained from (i) True Positive ($TP$) (when the real value is positive and prediction is positive); (ii) True Negative ($TN$) (when the real value is negative and the prediction is negative); (iii) False Positive ($FP$) (when the real values is negative but the prediction is positive) and (iv) False Negative ($FN$) (when the real value is positive but the prediction is negative).  Therefore  accuracy is described as the ratio of correctly classified instances, which is the total number of correct predictions divided by the total number of predictions made for a dataset; meanwhile,  precision and recall are used to quantify how well the proposed algorithm matches the ground truth. Precision quantifies the number of positive class predictions that actually belong to the positive class. In addition, Recall quantifies the number of positive class predictions made out of all positive examples in the dataset. Finally, the F1-score provides a way to combine both precision and recall into a single measure that captures both properties.

\begin{equation} 
	\begin{split}
&Accuracy =\frac{TP+TN}{TP+TN+FP+FN}\\ 
&Precision =\frac{TP}{TP+FP} \\
&Recall =\frac{TP}{TP+FN} \\
&F1-Score =2*\frac{Precision*Recall}{Precision+Recall} \\
\end{split}
\label{eq:ev:formulas}\end{equation}

\begin{figure}\centering
  \includegraphics[width=.6\textwidth]{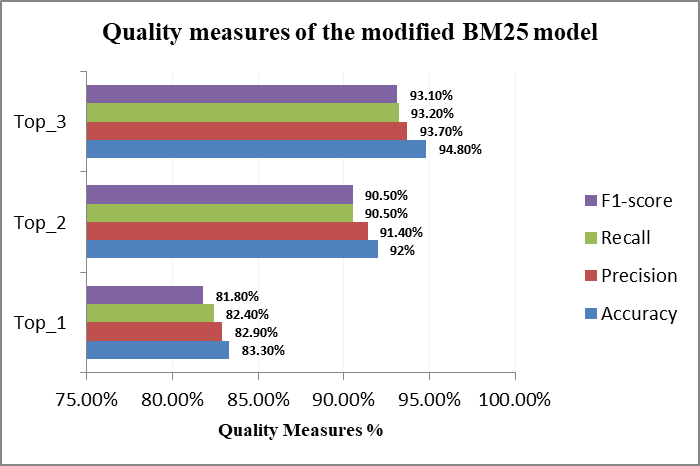}  
  \caption{Score-based Methods: Quality measures}
  \label{fig:QualityModifiedBM25}
\end{figure}

\autoref{tab:bm25} and~\autoref{fig:QualityModifiedBM25} summarizes the performance of our score-based approach, based on the BM25 model.

\begin{table}[h] \centering
\caption{Score-based Methods: Quality measures}
\begin{tabular}{cccc} \cline{1-4} \textbf{} &\textbf{$Top\_1$} &\textbf{$Top\_2$} &\textbf{$Top\_3$}\\ \hline
\textbf{Accuracy}& 0.833 & 0.920 & 0.948 \\ 
\textbf{Precision}& 0.829 & 0.914 & 0.937 \\ 
\textbf{Recall}& 0.824 & 0.905 & 0.932 \\ 
\textbf{F1-score}& 0.818 & 0.905 & 0.931 \\ \hline
\end{tabular}  \label{tab:bm25} \end{table}

We evaluate the performance  of our nearest neighbors models over reduced dimension dataset with 400, 500, 600, 700 and 800 principal components. KNN results are shown in~\autoref{tab:knn}. The results represent the best values for $Top\_1$: 700 principal components with Accuracy (0.8427). The best fit with $Top\_2$ is 600 principal components with Accuracy (0.8874). However, 500 principal components are also suitable for the F1-score (0.8241). In addition to the $Top\_3$, the exact result is 400 principal components with Accuracy (0.9201). On the other hand, 700 principal components are also suitable for the F1-score (0.8263). Regarding FKNN,~\autoref{tab:Fknn} illustrates that the best fit with $Top\_1$ for FKNN is 800 principal components with accuracy (0.8459). In addition, the best fit with $Top\_2$ is 600 principal components with accuracy (0.9153), also suitable for the F1-score (0.8302). In addition to the $Top\_3$, the best result is 600 principal components with accuracy (0.9413) and F1-score (0.8302).   We would like to draw attention to the values of $Top\_1$ results in~\autoref{tab:knn} and~\autoref{tab:Fknn}. Both of them are the same, which indicates that recall equals precision and accuracy. This means that the model is somehow "balanced", that is, its ability to correctly classify positive samples ($TP$+$FN$) is the same as its ability to correctly classify negative samples ($TN$+$FP$). Therefore, the F1-score is also the same~\cite{measure_parameter}.

\begin{table}[h] \centering
\caption{KNN: Quality measures by number of PCA components \label{tab:knn} }
\begin{tabular}{cccccc} \cline{1-6} \textbf{PCA} &\textbf{400} &\textbf{500} &\textbf{600}&\textbf{700} &\textbf{800}\\ \hline
\multicolumn{6}{c}{\textbf{$Top\_1$}}\\ \hline
\textbf{Accuracy}& 0.8339 & 0.8348	& 0.8367 & 0.8427 & 0.8332 \\
\textbf{Precision}& 0.8339 & 0.8348	& 0.8367 & 0.8427 & 0.8332  \\ 
\textbf{Recall}& 0.8339 & 0.8348 & 0.8367 & 0.8427 & 0.8332  \\ 
\textbf{F1-score}& 0.8339 & 0.8348	& 0.8367 & 0.8427 & 0.8332  \\ \hline
\multicolumn{6}{c}{\textbf{$Top\_2$}}\\ \hline
\textbf{Accuracy}& 0.8862 & 0.8871 & 0.8874 & 0.8871 & 0.8865 \\
\textbf{Precision}& 0.8405 & 0.8441 & 0.8403 & 0.8408 & 0.8346 \\ 
\textbf{Recall}& 0.8019 & 0.805 & 0.8072 & 0.8072 & 0.8063 \\ 
\textbf{F1-score}& 0.8207 & 0.8241 & 0.8234 & 0.8237 & 0.8202 \\ \hline
\multicolumn{6}{c}{\textbf{$Top\_3$}}\\ \hline
\textbf{Accuracy}& 0.9201 & 0.9125 & 0.9141 & 0.915 & 0.9137 \\
\textbf{Precision}& 0.8302 & 0.8313 & 0.8282 & 0.8048 & 0.8235\\ 
\textbf{Recall}& 0.78 & 0.7793 & 0.7873 & 0.7844 & 0.7841 \\ 
\textbf{F1-score}& 0.8043 & 0.8045 & 0.8072 & 0.8263 & 0.8033 \\ \hline
\end{tabular} \end{table}

\begin{table}[h] \centering
 \caption{FKNN: Quality measures by  PCA components \label{tab:Fknn}}
\begin{tabular}{cccccc} \cline{1-6} \textbf{PCA} &\textbf{400} &\textbf{500} &\textbf{600}&\textbf{700} &\textbf{800}\\ \hline
\multicolumn{6}{c}{\textbf{$Top\_1$}}\\ \hline
\textbf{Accuracy}& 0.8396 & 0.8424 & 0.8446 & 0.8443 & 0.8459 \\
\textbf{Precision}& 0.8396 & 0.8424 & 0.8446 & 0.8443 & 0.8459 \\ 
\textbf{Recall}& 0.8396 & 0.8424 & 0.8446 & 0.8443 & 0.8459 \\ 
\textbf{F1-score}& 0.8396 & 0.8424 & 0.8446 & 0.8443 & 0.8459 \\ \hline
\multicolumn{6}{c}{\textbf{$Top\_2$}}\\ \hline
\textbf{Accuracy}& 0.8957 & 0.9141 & 0.9153	& 0.9061 & 0.9096 \\
\textbf{Precision}& 0.7908 & 0.8289 & 0.8317 & 0.8165 & 0.8216 \\ 
\textbf{Recall}& 0.7885 & 0.8266 &	0.8288 & 0.8126 & 0.8177 \\ 
\textbf{F1-score}& 0.7897 & 0.8278 & 0.8302 & 0.8146 & 0.8196 \\ \hline
 \multicolumn{6}{c}{\textbf{$Top\_3$}}\\ \hline
\textbf{Accuracy}& 0.9315 & 0.94 & 0.9413 &	0.9362 & 0.9362 \\
\textbf{Precision}& 0.7949 & 0.8291 & 0.8319 & 0.8129 & 0.8167\\ 
\textbf{Recall}& 0.7803	& 0.8231 & 0.8285 & 0.8088 & 0.8123 \\ 
\textbf{F1-score}& 0.7875 & 0.8261 & 0.8302 & 0.8109 & 0.8145 \\ \hline
\end{tabular} \end{table}  

\autoref{tab:xgboost} presents the performance of our XGBoost classifier, which was also obtained for the principal components (400, 500, 600, 700 and 800), which declares selecting 400 as the best value for the principal component for $Top\_1$, $Top\_2$ and $Top\_3$; accuracy 0.7071 for $Top\_1$; 0.8126 for $Top\_2$ and 0.8568 for $Top\_3$.

\begin{table}[h] \centering\caption{XGBoost: Quality measures by  PCA components} \label{tab:xgboost}
\begin{tabular}{cccccc} \cline{1-6} \textbf{PCA} &\textbf{400} &\textbf{500} &\textbf{600}&\textbf{700} &\textbf{800}\\ \hline
\multicolumn{6}{c}{\textbf{$Top\_1$}}\\ \hline
\textbf{Accuracy}& 0.7071 & 0.6909 &0.6928 &0.6885 &0.6923 \\
\textbf{Precision}& 0.5908& 0.5651 &0.5664 &0.5395 &0.5648 \\ 
\textbf{Recall}&0.5722 &0.5467  & 0.5471&0.5176 &0.5411 \\ 
\textbf{F1-score}&0.5814 & 0.5557 &0.5577 &0.5283 & 0.5526\\ \hline
\multicolumn{6}{c}{\textbf{$Top\_2$}}\\ \hline
\textbf{Accuracy}&0.8126 & 0.8095 & 0.7949 & 0.7994&0.7946 \\
\textbf{Precision}&0.6850 & 0.6756 & 0.6445&0.6179 &0.6544 \\ 
\textbf{Recall}&0.6619 &0.6556  & 0.6348&0.6058 &0.6424 \\ 
\textbf{F1-score}&0.6733 & 0.6655 &0.6396 &0.6117 & 0.6483\\ \hline
 \multicolumn{6}{c}{\textbf{$Top\_3$}}\\ \hline
\textbf{Accuracy}&0.8568 & 0.8534 & 0.8404&0.8489 &0.8389 \\
\textbf{Precision}& 0.7243 &0.7059 &0.6878 &0.6391 &0.6647 \\ 
\textbf{Recall}& 0.7114 & 0.6987 &0.6723 &0.6272 &0.6597 \\ 
\textbf{F1-score}&0.7178 & 0.7023 &0.6799 &0.6331 &0.6622 \\ \hline
\end{tabular}  \end{table}

The performance of our MLP classifier was obtained also for  400, 500, 600, 700 and 800 principal components, as it is shown~\autoref{tab:MLP}, which suggests to select $800$ as the best value for the principal component for $Top\_1$, $Top\_2$ and $Top\_3$;  accuracy  0.8388 for $Top\_1$; 0.8436 for  $Top\_2$ and 0.8436 for  $Top\_3$.

\begin{table}[h] \centering\caption{MLP: Quality measures by  PCA components \label{tab:MLP}} 
\begin{tabular}{cccccc} \cline{1-6} \textbf{PCA} &\textbf{400} &\textbf{500} &\textbf{600}&\textbf{700} &\textbf{800}\\ \hline
\multicolumn{6}{c}{\textbf{$Top\_1$}}\\ \hline
\textbf{Accuracy}& 0.8327 & 0.8218 & 0.8366 & 0.8346 &	0.8388 \\
\textbf{Precision}& 0.8332 & 0.8220 & 0.8362 & 0.8343 & 0.8391 \\ 
\textbf{Recall}& 0.8324 & 0.8218 & 0.8360 & 0.8340 & 0.8389 \\ 
\textbf{F1-score}& 0.8328 & 0.8219 & 0.8361 & 0.8342 & 0.8390 \\ \hline
\multicolumn{6}{c}{\textbf{$Top\_2$}}\\ \hline
\textbf{Accuracy}& 0.8407 &	0.8324 & 0.8417 & 0.8404 & 0.8436 \\
\textbf{Precision}& 0.9688 & 0.9717 & 0.9814 & 0.9803 & 0.9815 \\ 
\textbf{Recall}& 0.8324 & 0.8218 & 0.8360 &	0.8340 & 0.8389 \\ 
\textbf{F1-score}& 0.8954 & 0.8905 & 0.9029 & 0.9013 & 0.9046 \\ \hline
 \multicolumn{6}{c}{\textbf{$Top\_3$}}\\ \hline
\textbf{Accuracy}& 0.8411 & 0.8327 & 0.8420	& 0.8407 & 0.8436\\
\textbf{Precision}& 0.9908 & 0.9907 & 0.9988 & 0.9950 & 0.9965 \\ 
\textbf{Recall}& 0.8324 & 0.8218 & 0.8360 & 0.8340 & 0.8389 \\ 
\textbf{F1-score}& 0.9047 & 0.8983 & 0.9102 & 0.9074 & 0.9109 \\ \hline
\end{tabular}  \end{table}

\begin{figure}[ht!]
    \includegraphics[width=0.5\textwidth]{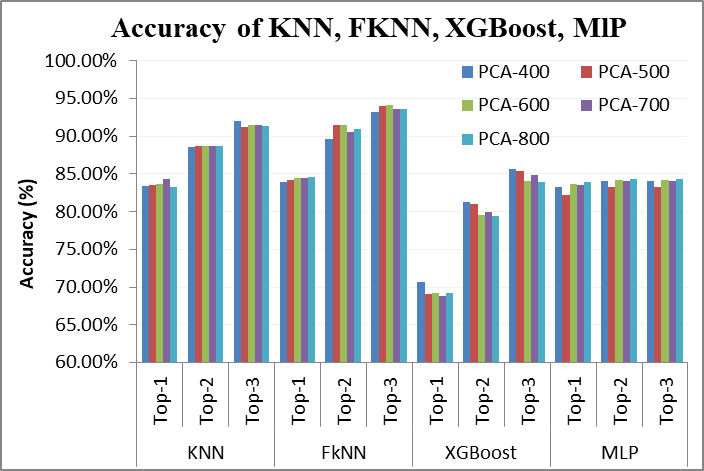}\hfill
    \includegraphics[width=0.5\textwidth]{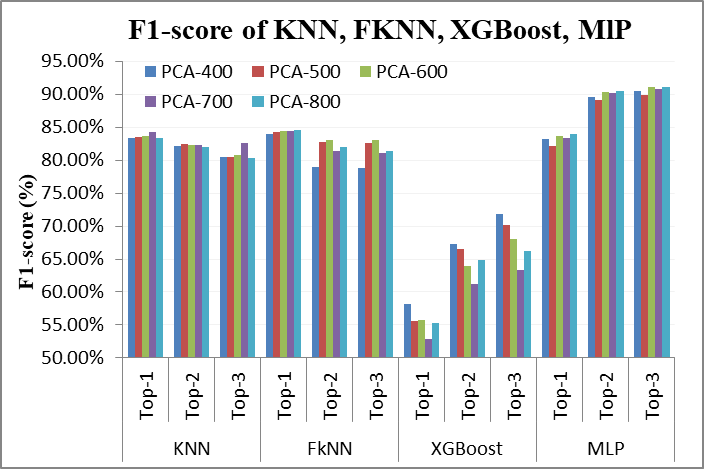}
\caption{Accuracy \& F1-score of KNN, FKNN, XGBoost, MLP.} \label{fig:Accuracy-F1-score-KNN-FKNN-MlP} 
\end{figure}

Additionally, we have assessed the results for the accuracy and the F1-score for the last four algorithms with respect to $Top\_1$, $Top\_2$ and  $Top\_3$ on a reduced dimension dataset with 400, 500, 600, 700 and 800 principal components; as show in~\autoref{fig:Accuracy-F1-score-KNN-FKNN-MlP}. The FKNN obtains the best accuracy on $Top\_1$, $Top\_2$ and $Top\_3$. The highest accuracy, in $Top\_1$ with the appropriate PCA which is PCA-800 is (84.59\%), and in $Top\_2$ and $Top\_3$ with the suitable PCA that's PCA-600 is (91.53\%, 94.13\%), respectively.  The best algorithm to obtain the highest result in the F1-score is FKNN in $Top\_1$ with the appropriate principal component which is PCA-800 at the rate of 84.59\%, while in $Top\_2$ and $Top\_3$, the best algorithm is MLP with PCA-800 at the rate of 90.46\%, 91.09\%, respectively.

\begin{figure}[h!] \centering
\color{black} \includegraphics[scale = 0.65]{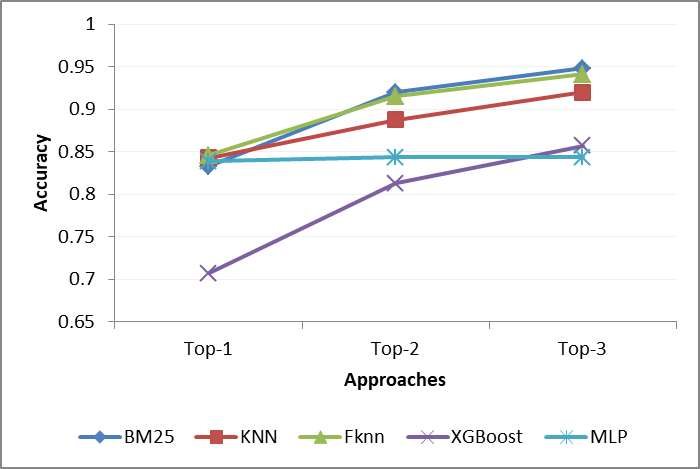} \caption{Comparative study between all algorithms.} \label{fig:all_alg} 
\end{figure}

To sum up,~\autoref{fig:all_alg} shows the comparison in terms of accuracy  among all the models (for their optimal values) regarding  $Top\_1$, $Top\_2$ and $Top\_3$.  The best in $Top\_1$ is FKNN followed by KNN. As well as, the appropriate one in $Top\_2$, $Top\_3$ is BM25 followed by FKNN.

\section{Conclusions}
\label{sec:conclusions} 
In this paper, we propose a solution for automatic classification of groceries in a digital transformation scenario, where the availability of quality data is key for retailers and costumers. 
Besides, food market is a very dynamic context where new products emerge daily, so product catalogs are composed of a huge amount of data that needs constant updates. Within this context,  the Spanish company {\em Midiadia} provides retailers with quality data about their product catalogs. {\em Midiadia} processes the information from the products packaging and labeling to offer relevant knowledge to retailers that eventually supports commercial processes (offers, customization, etc.). Our main objective is to collaborate with  {\em Midiadia} to improve the efficiency of one of their internal processes: classification of new products into the ontology the company uses. In particular, we focused on providing an automatic classification mechanism that replaces the manual task of deciding to which {\em Variety} (one of the taxonomy levels that is composed of $159$ values) a new product belongs.

With this aim, we worked with three different alternatives in order to decide the most appropriate: score-based algorithms, machine learning approaches, and deep neural networks. After an intensive work to preprocess the dataset (MDD-DS, composed by $20,888$ products), we applied the following algorithms. First, we define a score-based algorithm based on the probabilistic model BM25, but adapting the formulation to our specific problem of groceries. Then, we applied the KNN, FKNN and XGBoost algorithms, after reducing the data dimensionality with PCA. Finally, we also applied an MLP algorithm using the same PCA mechanism for reducing data dimensionality. 

The main objective is providing a totally automatic classification tool that directly offers one result: the most appropriate {\em Variety} for a new product ($Top\_1$ option). However, and considering the company benefits, we also studied other two alternatives: (i) $Top\_2$ that obtains the 2 most appropriate {\em Varieties} for a new product and (ii) $Top\_3$ that obtains the 3 most appropriate {\em Varieties} for a new product. Both of them would also help the classification process by turning the problem of classifying a new product within $159$  categories into a problem of classifying a new product within $2$ or $3$ categories. According to our results, the best in $Top\_1$ is FKNN, closely followed by KNN. Although in $Top\_2$ and $Top\_3$ the score-based algorithm performs better, closely followed by FKNN.

We are currently working on the creation of a dictionary of synonyms for ingredients. This need arises because the current regulations allow manufacturers to include different terms for the same element (vitamin C or ascorbic acid, for instance). Additionally, we are working on extending our analysis to other approaches, such as Convolutional Neural Network (CNN), Long Short-Term Memory (LSTM), Deep Bidirectional LSTMs (BILSTM), Bidirectional Encoder Representations from Transformers (BERT) and Support Vector Machine (SVM). Regarding the course of dimensionality, applying DNN with abstract features is a natural next step provided that the issues related with memory consumption during training can be mitigated.  After that, we expect to perform a comparison experiment with deep neural networks using additional hidden layers. Besides, the research work already done for the classification problem is the basis of a recommender algorithm to provide product alternatives to the customer when the desired product is not available. With this aim, we are currently working on a multidimensional recommender that takes into account the list of ingredients and other relevant data, such as calories, allergens, healthy aspects, etc.

\section{acknowledgements}

The authors would like to thank the European Regional Development Fund (ERDF) and the Galician Regional 7 Government, under the agreement for funding the Atlantic Research Center for Information and Communication Technologies (atlanTTIC), and the Spanish Ministry of Economy and Competitiveness, under the National Science Program (TEC2017-84197-C4-2-R).

\appendix

\section{Extra Tables}

\begin{table}[H] \centering
\caption{Definition of distance measures~\cite{abu2019effects,distance}}
\begin{tabular}{p{190pt}| p{190pt}}
 \hline
\textbf{Cityblock}, also known as Manhattan distance, It examines the absolute differences between the opposite values in vectors as shown in~\autoref{eq:ed}.
\begin{equation} CB[a,b] =\sum_{i = 1}^ n |a_i-b_i| \label{eq:ed}\end{equation} 
 &
\textbf{Cosine} is a measure of similarity between two nonzero vectors of an internal product space that measures the cosine of the angle between them as shown in~\autoref{eq:co}.
\begin{equation} cos(a,b) = \frac {a \cdot b}{||a|| \cdot ||b||} \label{eq:co}\end{equation} \\\hline

\textbf{Correlation} is a measure of dependency between two paired random vectors of arbitrary dimension, not necessarily equal as shown in~\autoref{eq:cor}. 
\begin{equation} CorD[a,b] =\frac{{}\sum_{i=1}^{n} (a_i - \overline{a})(b_i - \overline{b})}
{\sqrt{\sum_{i=1}^{n} (a_i - \overline{a})^2(b_i - \overline{b})^2}} \label{eq:cor}\end{equation} 
&
\textbf{Euclidean} is the ordinary straight-line distance between two points in Euclidean space as shown in~\autoref{eq:eud}. 
\begin{equation} ED[a,b] =\sqrt{\sum_{i = 1}^ n |a_i-b_i|^2} \label{eq:eud}\end{equation} \\\hline

\textbf{Seuclidean}, Standardized Euclidean distance. Defined as the Euclidean distance calculated on standardized data as shown in~\autoref{eq:seuc}. \begin{equation} SED[a,b] =\sqrt{\sum_{i=1}^n\frac{1}{k_i^2}(a_i-b_i)^2} \label{eq:seuc}\end{equation}
&
\textbf{Jaccard} distance measures the difference between sets of samples, it is a complement to the Jaccard similarity coefficient and is obtained by subtracting the Jaccard coefficient from one as shown in~\autoref{eq:jac}.
\begin{equation} Jac[a,b] =\frac{\sum_{i}\min(a_i,b_i)}{\sum_{i}\max(a_i,b_i)} \label{eq:jac}\end{equation} \\\hline

\textbf{Hamming}, the percentage of coordinates that differ as shown in~\autoref{eq:ham}.
\begin{equation} HD(a,b) =\sum _{i=1}^n 1_{a_i\neq b_i} \label{eq:ham}\end{equation} 
&
\textbf{Chebychev}, Maximum coordinate difference as shown in~\autoref{eq:cheb}. \begin{equation} CD[a,b] =\max_i|a_i-b_i| \label{eq:cheb}\end{equation} \\\hline
\textbf{Spearman's $\rho$} is a nonparametric measure of the statistical dependence of rank correlation between the rankings of two variables. Evaluate how well the relationship between two variables can be described using a monotonic function as shown in~\autoref{eq:sp}.
&
\begin{equation} \rho = 1- {\frac {6 \sum d_i^2}{n(n^2 - 1)}} \label{eq:sp}\end{equation} 
 where $d =$ the pairwise distances of the ranks of the variables $a_i\ and\ b_i.\ n =$ the number of samples.\\\hline
\end{tabular}  \label{tab:distance_measure} \end{table} 


\bibliographystyle{ieeetr}
\bibliography{bibliography}

\begin{thebibliography}{10}

\bibitem{REINARTZ2019350}
W.~Reinartz, N.~Wiegand, and M.~Imschloss, ``The impact of digital transformation on the retailing value chain,'' {\em International Journal of Research in Marketing}, vol.~36, no.~3, pp.~350 -- 366, 2019.

\bibitem{wessel2020unpacking}
L.~Wessel, A.~Baiyere, R.~Ologeanu-Taddei, J.~Cha, and T.~Jensen, ``Unpacking the difference between digital transformation and it-enabled organizational transformation,'' {\em Journal of Association of Information Systems}, 2020.

\bibitem{Nielsen-FMI}
Nielsen, ``The digitally engaged food shopper: Developing your omnichannel collaboration model,'' 2018.

\bibitem{bahn2020descriptive}
R.~A. Bahn and G.~K. Abebe, ``A descriptive analysis of food retailing in lebanon: Evidence from a cross-sectional survey of food retailers,'' in {\em Food Supply Chains in Cities}, pp.~289--346, Springer, 2020.

\bibitem{hafez2018comparative}
M.~M. Hafez, R.~P.~D. Redondo, and A.~F. Vilas, ``A comparative performance study of na{\"\i}ve and ensemble algorithms for e-commerce,'' in {\em 2018 14th International Computer Engineering Conference (ICENCO)}, pp.~26--31, IEEE, 2018.

\bibitem{eu_law}
E.~Commission, ``{General Food Law}.'' \url{https://ec.europa.eu/food/safety/general\_food\_law\_en}, 2002.
\newblock [Online; accessed 2019].

\bibitem{es_law}
BOE, ``{Real Decreto Legislativo 1/2007, de 16 de noviembre, por el que se aprueba el texto refundido de la Ley General para la Defensa de los Consumidores y Usuarios y otras leyes complementarias.}.'' \url{https://www.boe.es/eli/es/rdlg/2007/11/16/1/con}, 30-11-2007.
\newblock [Online; accessed 2019].

\bibitem{gl_law}
BOE, ``{Ley 2/2012, de 28 de marzo, gallega de protección general de las personas consumidoras y usuarias.}.'' \url{https://www.boe.es/eli/es-ga/l/2012/03/28/2}, 27-April-2012.
\newblock [Online; accessed 2019].

\bibitem{baz2016context}
I.~Baz, E.~Yoruk, and M.~Cetin, ``Context-aware hybrid classification system for fine-grained retail product recognition,'' in {\em 2016 IEEE 12th Image, Video, and Multidimensional Signal Processing Workshop (IVMSP)}, pp.~1--5, IEEE, 2016.

\bibitem{fuchs2019towards}
K.~Fuchs, T.~Grundmann, and E.~Fleisch, ``Towards identification of packaged products via computer vision: Convolutional neural networks for object detection and image classification in retail environments,'' in {\em Proceedings of the 9th International Conference on the Internet of Things}, pp.~1--8, 2019.

\bibitem{hafez2016effective}
M.~M. Hafez, M.~E. Shehab, E.~El~Fakharany, {\em et~al.}, ``Effective selection of machine learning algorithms for big data analytics using apache spark,'' in {\em International Conference on Advanced Intelligent Systems and Informatics}, pp.~692--704, Springer, 2016.

\bibitem{peng2020rp2k}
J.~Peng, C.~Xiao, X.~Wei, and Y.~Li, ``Rp2k: A large-scale retail product dataset forfine-grained image classification,'' {\em arXiv preprint arXiv:2006.12634}, 2020.

\bibitem{baz2019statistical}
{\.I}.~Baz, {\em Statistical methods for fine-grained retail product recognition}.
\newblock PhD thesis, 2019.

\bibitem{bianchi2020retail}
T.~Bianchi-Aguiar, A.~H{\"u}bner, M.~A. Carravilla, and J.~F. Oliveira, ``Retail shelf space planning problems: A comprehensive review and classification framework,'' {\em European Journal of Operational Research}, 2020.

\bibitem{goldman2019precise}
E.~Goldman, R.~Herzig, A.~Eisenschtat, J.~Goldberger, and T.~Hassner, ``Precise detection in densely packed scenes,'' in {\em Proceedings of the IEEE Conference on Computer Vision and Pattern Recognition}, pp.~5227--5236, 2019.

\bibitem{gundimeda2019automated}
V.~Gundimeda, R.~S. Murali, R.~Joseph, and N.~N. Babu, ``An automated computer vision system for extraction of retail food product metadata,'' in {\em First International Conference on Artificial Intelligence and Cognitive Computing}, pp.~199--216, Springer, 2019.

\bibitem{wang2016matching}
X.~Wang, Z.~Sun, W.~Zhang, Y.~Zhou, and Y.-G. Jiang, ``Matching user photos to online products with robust deep features,'' in {\em Proceedings of the 2016 ACM on International Conference on Multimedia Retrieval}, pp.~7--14, 2016.

\bibitem{zhong2020temporal}
C.~Zhong, L.~Jiang, Y.~Liang, H.~Sun, and C.~Ma, ``Temporal multiple-convolutional network for commodity classification of online retail platform data,'' in {\em Proceedings of the 2020 12th International Conference on Machine Learning and Computing}, pp.~236--241, 2020.

\bibitem{pobbathi2020automated}
N.~R. Pobbathi, A.~Dong, and Y.~Chang, ``Automated categorization of products in a merchant catalog,'' Jan.~7 2020.
\newblock US Patent 10,528,907.

\bibitem{seth2020method}
S.~Seth, B.~S. Johnson, R.~Kennedy, and N.~Kothari, ``Method and system to categorize items automatically,'' July~7 2020.
\newblock US Patent 10,706,076.

\bibitem{ecommerce_auto}
V.~Gupta, H.~Karnick, A.~Bansal, and P.~Jhala, ``Product classification in e-commerce using distributional semantics,'' {\em CoRR}, vol.~abs/1606.06083, 2016.

\bibitem{baeza}
R.~Baeza-Yates and B.~Ribeiro-Neto, {\em Modern Information Retrieval: The Concepts and Technology behind Search}.
\newblock USA: Addison-Wesley Publishing Company, 2nd~ed., 2011.

\bibitem{chen2020shape}
H.~Chen, J.~Pang, M.~Koo, and V.~M. Patrick, ``Shape matters: Package shape informs brand status categorization and brand choice,'' {\em Journal of Retailing}, vol.~96, no.~2, pp.~266--281, 2020.

\bibitem{wei2020deep}
Y.~Wei, S.~Tran, S.~Xu, B.~Kang, and M.~Springer, ``Deep learning for retail product recognition: challenges and techniques,'' {\em Computational Intelligence and Neuroscience}, vol.~2020, 2020.

\bibitem{wei2019deep}
X.-S. Wei, J.~Wu, and Q.~Cui, ``Deep learning for fine-grained image analysis: A survey,'' {\em arXiv preprint arXiv:1907.03069}, 2019.

\bibitem{dashtipour2020hybrid}
K.~Dashtipour, M.~Gogate, J.~Li, F.~Jiang, B.~Kong, and A.~Hussain, ``A hybrid persian sentiment analysis framework: Integrating dependency grammar based rules and deep neural networks,'' {\em Neurocomputing}, vol.~380, pp.~1--10, 2020.

\bibitem{customer_preferences}
P.~Korgaonkar, R.~Silverblatt, and T.~Girard, ``Online retailing, product classifications, and consumer preferences,'' {\em Internet Research}, vol.~16, no.~3, pp.~267--288, 2006.

\bibitem{customer_behaviour}
Z.~V. Ravnik~R., Solina~F., ``Modelling in-store consumer behaviour using machine learning and digital signage audience measurement data,'' in {\em Video Analytics for Audience Measurement. VAAM 2014}, pp.~691--695, Springer, 2014.

\bibitem{holy2017clustering}
V.~Hol{\`y}, O.~Sokol, and M.~{\v{C}}ern{\`y}, ``Clustering retail products based on customer behaviour,'' {\em Applied Soft Computing}, vol.~60, pp.~752--762, 2017.

\bibitem{okapi_exp}
S.~Robertson, S.~Walker, and M.~Beaulieu, ``Experimentation as a way of life: Okapi at trec,'' {\em Information Processing \& Management}, vol.~36, no.~1, pp.~95 -- 108, 2000.

\bibitem{abu2019effects}
H.~A. Abu~Alfeilat, A.~B. Hassanat, O.~Lasassmeh, A.~S. Tarawneh, M.~B. Alhasanat, H.~S. Eyal~Salman, and V.~S. Prasath, ``Effects of distance measure choice on k-nearest neighbor classifier performance: A review,'' {\em Big data}, vol.~7, no.~4, pp.~221--248, 2019.

\bibitem{keller1985fuzzy}
J.~M. Keller, M.~R. Gray, and J.~A. Givens, ``A fuzzy k-nearest neighbor algorithm,'' {\em IEEE transactions on systems, man, and cybernetics}, no.~4, pp.~580--585, 1985.

\bibitem{derrac2014fuzzy}
J.~Derrac, S.~Garc{\'\i}a, and F.~Herrera, ``Fuzzy nearest neighbor algorithms: Taxonomy, experimental analysis and prospects,'' {\em Information Sciences}, vol.~260, pp.~98--119, 2014.

\bibitem{wankhede2014analytical}
S.~B. Wankhede, ``Analytical study of neural network techniques: Som, mlp and classifier-a survey,'' {\em IOSR Journal of Computer Engineering}, vol.~16, no.~3, pp.~86--92, 2014.

\bibitem{al2013lemmatizing}
E.~T. Al-Shammari, ``Lemmatizing, stemming, and query expansion method and system,'' June~25 2013.
\newblock US Patent 8,473,279.

\bibitem{wang2020principal}
D.~Wang and J.~Xu, ``Principal component analysis in the local differential privacy model,'' {\em Theoretical Computer Science}, vol.~809, pp.~296--312, 2020.

\bibitem{balakrishnama1998linear}
S.~Balakrishnama and A.~Ganapathiraju, ``Linear discriminant analysis-a brief tutorial,'' in {\em Institute for Signal and information Processing}, vol.~18, pp.~1--8, 1998.

\bibitem{kunang2018automatic}
Y.~N. Kunang, S.~Nurmaini, D.~Stiawan, A.~Zarkasi, {\em et~al.}, ``Automatic features extraction using autoencoder in intrusion detection system,'' in {\em 2018 International Conference on Electrical Engineering and Computer Science (ICECOS)}, pp.~219--224, IEEE, 2018.

\bibitem{wang2016auto}
Y.~Wang, H.~Yao, and S.~Zhao, ``Auto-encoder based dimensionality reduction,'' {\em Neurocomputing}, vol.~184, pp.~232--242, 2016.

\bibitem{jung2003principal}
Y.-M. Jung, ``Principal component analysis based two-dimensional (pca-2d) correlation spectroscopy: Pca denoising for 2d correlation spectroscopy,'' {\em Bulletin of the Korean Chemical Society}, vol.~24, no.~9, pp.~1345--1350, 2003.

\bibitem{reddy2020analysis}
G.~T. Reddy, M.~P.~K. Reddy, K.~Lakshmanna, R.~Kaluri, D.~S. Rajput, G.~Srivastava, and T.~Baker, ``Analysis of dimensionality reduction techniques on big data,'' {\em IEEE Access}, vol.~8, pp.~54776--54788, 2020.

\bibitem{bm25_trec5}
M.~M. Beaulieu, M.~Gatford, X.~Huang, S.~Robertson, S.~Walker, and P.~Williams, ``Okapi at trec-5,'' in {\em The Fifth Text REtrieval Conference (TREC-5)}, pp.~143--165, Gaithersburg, MD: NIST, January 1997.

\bibitem{rastin2020generalized}
N.~Rastin, M.~Z. Jahromi, and M.~Taheri, ``A generalized weighted distance k-nearest neighbor for multi-label problems,'' {\em Pattern Recognition}, p.~107526, 2020.

\bibitem{hassanat2014solving}
A.~B. Hassanat, M.~A. Abbadi, G.~A. Altarawneh, and A.~A. Alhasanat, ``Solving the problem of the k parameter in the knn classifier using an ensemble learning approach,'' {\em arXiv preprint arXiv:1409.0919}, 2014.

\bibitem{thongsuwan2021convxgb}
S.~Thongsuwan, S.~Jaiyen, A.~Padcharoen, and P.~Agarwal, ``Convxgb: A new deep learning model for classification problems based on cnn and xgboost,'' {\em Nuclear Engineering and Technology}, vol.~53, no.~2, pp.~522--531, 2021.

\bibitem{shilong2021machine}
Z.~Shilong {\em et~al.}, ``Machine learning model for sales forecasting by using xgboost,'' in {\em 2021 IEEE International Conference on Consumer Electronics and Computer Engineering (ICCECE)}, pp.~480--483, IEEE, 2021.

\bibitem{feng2020multi}
X.~Feng, G.~Ma, S.-F. Su, C.~Huang, M.~K. Boswell, and P.~Xue, ``A multi-layer perceptron approach for accelerated wave forecasting in lake michigan,'' {\em Ocean Engineering}, vol.~211, p.~107526, 2020.

\bibitem{measure_parameter}
J.~Brownlee, ``{How to Calculate Precision, Recall, and F-Measure for Imbalanced Classification}.'' \url{https://machinelearningmastery.com/precision-recall-and-f-measure-for-imbalanced-classification/}, 03-January-2020.
\newblock [Online; accessed 11-March-2020].

\bibitem{distance}
J.~Banda {\em et~al.}, {\em Framework for creating large-scale content-based image retrieval system (CBIR) for solar data analysis}.
\newblock PhD thesis, Montana State University-Bozeman, College of Engineering, 2011.

\end{thebibliography}

\end{document}